\documentclass{ifacconf}

\usepackage{graphicx,xcolor} 
\usepackage{amsmath} 
\usepackage{amssymb}  
\usepackage{subcaption}
\usepackage{comment}
\usepackage{algorithm}
\usepackage{algpseudocode}
\usepackage{natbib}
\usepackage{soul}
\usepackage{upgreek}

\usepackage[hyphens]{url}
\usepackage{breakurl}

\usepackage{tikz}
\usetikzlibrary{shapes.geometric, arrows, positioning, arrows.meta}

\tikzstyle{process} = [rectangle, rounded corners, minimum width=1cm, minimum height=0.75cm, 
text centered, draw=black, fill=white]
\tikzstyle{arrow} = [line width=1mm,->,>=stealth, blue!20]

\usepackage{ctable}
\newcommand{\real}[1][]{\mathbb{R}^{#1}}                                
\newcommand{\nat}[1][]{\mathbb{N}^{#1}}                                 

\newcommand{\defeq}{:=}                                                 
\newcommand{\msub}[1]{_\mathrm{#1}}                                     
\newcommand{\msup}[1]{^\mathrm{#1}}                                     

\newcommand{\eye}[1]{\boldsymbol{I}_{#1}}                                          
\newcommand{\myfracB}[2]{{#1}/{#2}}                                     

\newcommand{\innerp}[2]{\left\langle {#1}, {#2} \right\rangle}     		

\newcommand{\clint}[2]{\left[#1, #2\right]}                             
\newcommand{\lopint}[2]{(#1, #2]}                                       

\renewcommand{\leq}{\leqslant}                                          
\renewcommand{\geq}{\geqslant}                                          

\newcommand{\intset}[1]{\{#1\}}

\def\unif{\mathbb{U}}

\newcommand{\mydef}[1]{{\textit{#1}}}

\newcommand{\eqnnt}[1]{\hyperref[#1]{(\ref*{#1})}}
\newcommand{\eqnsnt}[2]{\hyperref[#1]{(\ref*{#1})}
	and~\hyperref[#2]{(\ref*{#2})}}
\newcommand{\eqnsernt}[2]{\hyperref[#1]{(\ref*{#1})}--\hyperref[#2]{(\ref*{#2})}}

\newcommand{\eqn}[1]{\hyperref[#1]{Eqn.~(\ref*{#1})}}
\newcommand{\eqns}[2]{\hyperref[#1]{Eqns.~(\ref*{#1})} and~\hyperref[#2]{(\ref*{#2})}}
\newcommand{\eqnser}[2]{\hyperref[#1]{Eqns.~(\ref*{#1})}--\hyperref[#2]{(\ref*{#2})}}
\newcommand{\eqnf}[1]{\hyperref[#1]{Equation~(\ref*{#1})}}
\newcommand{\eqnfs}[2]{\hyperref[#1]{Equations~(\ref*{#1})} and~\hyperref[#2]{(\ref*{#2})}}

\newcommand{\scn}[1]{\hyperref[#1]{\S\ref*{#1}}}
\newcommand{\scns}[2]{\hyperref[#1]{\S\ref*{#1}} and~\hyperref[#2]{\ref*{#2}}}
\newcommand{\scnser}[2]{\hyperref[#1]{\S\ref*{#1}}--\hyperref[#2]{\ref*{#2}}}

\newcommand{\fig}[1]{\hyperref[#1]{Fig.~\ref*{#1}}}
\newcommand{\figs}[2]{\hyperref[#1]{Figs.~\ref*{#1}} and~\hyperref[#2]{\ref*{#2}}}
\newcommand{\figser}[2]{\hyperref[#1]{Figs.~\ref*{#1}}--\hyperref[#2]{\ref*{#2}}}
\newcommand{\figf}[1]{\hyperref[#1]{Figure~\ref*{#1}}}
\newcommand{\figfs}[2]{\hyperref[#1]{Figures~\ref*{#1}} and~\hyperref[#2]{\ref*{#2}}}
\newcommand{\figfser}[2]{\hyperref[#1]{Figures~\ref*{#1}}--\hyperref[#2]{\ref*{#2}}}

\newcommand{\tbl}[1]{\hyperref[#1]{Table~\ref*{#1}}}
\newcommand{\tbls}[2]{\hyperref[#1]{Tables~\ref*{#1}} and~\hyperref[#2]{\ref*{#2}}}
\newcommand{\tblser}[2]{\hyperref[#1]{Tables~\ref*{#1}}--\hyperref[#2]{\ref*{#2}}}

\newcommand{\apx}[1]{\hyperref[#1]{Appendix~\ref*{#1}}}

\newcommand{\chp}[1]{\hyperref[#1]{Chapter~\ref*{#1}}}
\newcommand{\chps}[2]{\hyperref[#1]{Chapters~\ref*{#1}} and~\hyperref[#2]{(\ref*{#2})}}
\newcommand{\chpser}[2]{\hyperref[#1]{Chapters~\ref*{#1}}--\hyperref[#2]{(\ref*{#2})}}

\newcommand{\prb}[1]{\hyperref[#1]{Problem~\ref*{#1}}}
\newcommand{\prp}[1]{\hyperref[#1]{Prop.~\ref*{#1}}}
\newcommand{\prpf}[1]{\hyperref[#1]{Proposition~\ref*{#1}}}

\newcommand{\corref}[1]{\hyperref[#1]{Cor.~\ref*{#1}}}
\newcommand{\algoref}[1]{\hyperref[#1]{Algorithm~\ref*{#1}}}
\newcommand{\asmref}[1]{\hyperref[#1]{Assumption~\ref*{#1}}}

\newcommand{\thmref}[1]{\hyperref[#1]{Theorem~\ref*{#1}}}
\newcommand{\thmsref}[2]{\hyperref[#1]{Theorems~\ref*{#1}} and~\hyperref[#2]{\ref*{#2}}}
\newcommand{\thmserref}[2]{\hyperref[#1]{Theorems~\ref*{#1}}--\hyperref[#2]{\ref*{#2}}}

\newcommand{\lemref}[1]{\hyperref[#1]{Lemma~\ref*{#1}}}

\newcommand{\algline}[1]{\hyperref[#1]{Line~\ref*{#1}}}
\newcommand{\alglines}[2]{\hyperref[#1]{Lines~\ref*{#1}} and~\hyperref[#2]{\ref*{#2}}}
\newcommand{\alglineser}[2]{\hyperref[#1]{Lines~\ref*{#1}}--\hyperref[#2]{\ref*{#2}}}

\def\nGridPts{N\msub{G}}
\def\nTimeSteps{N\msub{T}}
\def\nAction{|\mdpActionSet|}

\newcommand{\xCoord}[1][]{x_{#1}}
\newcommand{\xCoordGoal}{\bar{x}}
\newcommand{\tCoord}[1][]{t_{#1}}

\def\timeStep{\delta_t}
\def\threat{c}

\def\mdpStateSet{\mathcal{S}}
\def\mdpState{s}
\def\mdpActionSet{\mathcal{U}}
\def\mdpAction{u}
\def\mdpTransitionFcn{\uptau}
\def\rewardFcn{r}
\def\discount{\gamma}

\def\gridPath{\boldsymbol{s}}
\def\gridPathGen{\overline{\boldsymbol{s}}}

\newcommand{\gridPathLength}[1][]{L_{#1}}

\def\policy{\pi}
\newcommand{\valueFcn}[1][\policy]{V^{#1}}

\def\trainingFeatEx{\mu_\trainingDataset}
\def\aggregateFeature{\Phi}
\def\featError{\delta_\mu}

\def\nData{N}
\def\nGen{K}

\def\trainingDataset{\mathcal{Z}}
\def\genDataset{\overline{\mathcal{Z}}}

\newcommand{\QFcn}[1][\policy]{Q^{#1}}
\def\optimalQFcn{Q^*}

\def\featureFcn{\phi}

\newcommand{\featEx}[1][]{\mu^{#1}}
\newcommand{\featExTmp}{\mu}

\def\wtReward{w}
\def\wtQFcn{\theta}
\def\wtQFcnP{\theta'}

\def\IRLErrorThreshold{e_\mu}
\def\IRLMaxIterations{M\msub{I}}
\def\IRLLearningRate{\eta_I}
\def\IRLMaxPathIter{M\msub{P}}
\def\DQLLearningRate{\eta_Q}
\def\DQLLearningRateP{\eta_{Q'}}

\def\QLMaxEpisodes{M\msub{E}}
\def\QLMaxIterations{M\msub{Q}}

\def\selectionProb{\varepsilon}
\def\selectionProbA{\varepsilon_0}
\def\selectionProbB{\varepsilon_1}
\def\decayRate{d}

\def\rewardPlus{\ell}
\def\rewardPlusVec{\vec{\ell}}
\def\lossThreshold{\bar{\ell}}

\begin{document}

\begin{frontmatter}
	
	\title{Inverse Reinforcement Learning \\ for Minimum-Exposure Paths 
		in Spatiotemporally Varying Scalar Fields\thanksref{footnoteinfo}} 
	
	\thanks[footnoteinfo]{This research was sponsored by the DEVCOM Analysis Center and was 
	accomplished under 
		IDIQ Number W911NF-22-2-0001. The views and conclusions contained 
		in this document are those of the authors and should not be interpreted as representing
		the official policies, either expressed or implied, of the Army Research Office 
		or the U.S. Government. The U.S. Government is authorized to reproduce and distribute
		reprints for Government purposes notwithstanding any copyright notation herein.}
	
	\author[First]{Alex Ballentine} \quad\quad 
	\author[First]{Raghvendra V. Cowlagi} 
	
	\address[First]{Aerospace Engineering Department, 
		Worcester Polytechnic Institute, 
		Worcester, MA USA 01609  E-mail: aeballentine, rvcowlagi@ wpi.edu.}

	\begin{abstract}                
		Performance and reliability analyses of autonomous vehicles (AVs) can benefit
		from tools that ``amplify'' small datasets to synthesize larger volumes of
		plausible samples of the AV's behavior. We consider a specific instance of
		this data synthesis problem that addresses minimizing the AV's exposure to
		adverse environmental conditions during travel to a fixed goal location.
		The environment is characterized by a threat field, which is a 
		strictly positive scalar field with higher intensities corresponding to 
		hazardous and unfavorable conditions for the AV. We address the
		problem of synthesizing datasets of minimum exposure paths that resemble a
		training dataset of such paths. The main contribution of this paper is an inverse
		reinforcement learning (IRL) model to solve this problem. We consider
		time-invariant (static) as well as time-varying (dynamic) threat fields. We
		find that the proposed IRL model provides excellent performance in
		synthesizing paths from initial conditions not seen in the training dataset,
		when the threat field is the same as that used for training. Furthermore, we
		evaluate model performance on unseen threat fields and find low error in that
		case as well. Finally, we demonstrate the model's ability to synthesize
		distinct datasets when trained on different datasets with distinct
		characteristics.
	\end{abstract}
	
	
\end{frontmatter}

\section{Introduction}

Performance- and reliability analyses of an autonomous vehicle (AV) require an
understanding or ``explanation'' of the underlying autonomy stack, as well as a
large number of samples of the vehicle's operation \citep{Young2017}. However,
such information is difficult to obtain because of high cost of real-world
operations and proprietary hardware and software. Due to the widespread use of
machine learning in autonomy, an explanation of autonomous behaviors is not
always available, even to the manufacturer. More commonly, a high-level
description of the autonomy stack and a dataset of a small number of samples of
the AV's operation are available. In this situation, performance- and
reliability analyses can benefit on tools that can ``amplify'' the small dataset
to synthesize larger volumes of plausible samples of the AV's behavior.

Synthesis of a large volume of data that is statistically similar to a given
training dataset can be accomplished by a generative machine learning
model~\citep{Ruthotto2021}. Widely used generative models include generative
adversarial networks, variational autoencoders, and transformer neural networks.
In the present context, however, we seek an explainable model, as well as the
ability to incorporate a high-level description of the objectives of the
vehicle's autonomous behavior. To this end, we study inverse reinforcement
learning (IRL) for the specific problem of autonomous navigation in unstructured
environments.

\paragraph*{Related work:}
Imitation learning, or apprenticeship learning, aims to mimic an expert's
behavior. In imitation learning, the goal may either be to directly learn the
policy or to learn a reward function which explains an expert's
actions~\citep{naranjo-campos_expert_2024}. Two main approaches include
behavioral cloning (BC), following the direct approach, and IRL, 
following the indirect approach. 
BC maps the
environmental states to expert actions to decide new actions. However, BC relies
on previous demonstrations, making it less effective at responding to novel
situations~\citep{zare_survey_2024}. IRL aims to solve the challenge of
determining an agent's reward function from a given policy or observed
behavior~\citep{arora_survey_2021}.  Once the algorithm determines the reward
function, IRL algorithms use reinforcement learning (RL) to extrapolate the
policy from the reward function. IRL agents learn through interacting with the
environment and altering their behavior based on the results of their
actions~\citep{zare_survey_2024}.

IRL is most useful in applications where an explicit reward function is either
difficult to formulate or unknown, for instance in mimicking human behavior.
Additionally, IRL provides a more flexible framework which is capable of
adapting in dynamic environments, compared to RL where the reward function is
pre-defined and error-prone \citep{naranjo-campos_expert_2024}. Some
applications of IRL include training an agent to play video games
\citep{tucker_inverse_2018}, modeling human interactions and navigation
\citep{martinez-gil_using_2020}, and water management
\citep{likmeta_dealing_2021}. IRL has also been used for path-planning by
learning from expert demonstrations. Some examples include training algorithms
for self-driving cars \citep{huang_conditional_2023} and modeling navigation
through human-centric environments, such as crowds, where the shortest path is
less important \citep{feng_safe_2024}.

\citet{ng_algorithms_2000} proposed some of the first inverse reinforcement
algorithms by formalizing the IRL problem with Markov decision processes. The
basis for RL is that the reward function is the most representative definition
of a given task. IRL extends this to apprenticeship learning, where the reward
function is not known, but there are known expert demonstrations. At the time of
their work, they describe some of the primary problems with IRL, including
dealing with sub-optimal data and observations, learning "easy" reward
functions, partially observable environments, and maximizing the identifiability
of the reward function \citep{ng_algorithms_2000}. \citet{ziebart_maximum_2008}
propose maximum entropy approach to IRL that can provide a convex,
computationally efficient problem formulation.

One important challenge of IRL is the ambiguity of any potential solution. IRL
has always been an ill-posed problem, i.e., that there are multiple possible
reward functions that can describe the behavior of an expert
\citep{cao_identifiability_2021}. Some explored solutions include comparing the
value function of the expert with the value function of the learner, in which
case the error goes to zero if both policies are equivalent or both are optimal.
Another solution includes comparing the state-action pairs of the expert and
those of the learner \citep{arora_survey_2021}.

While the exploration inherent in IRL algorithms provides robustness compared to
BC, IRL is more computationally expensive to train. Further, IRL algorithms can
be unpredictable while training, making them dangerous to train outside of
simulation, and have poor sampling efficiency as a typical method involves
alternating between updating the reward and training the policy
\citep{zare_survey_2024}. \citet{brown_machine_2019} propose a method of
reducing the requisite number of demonstrations for IRL, called machine
teaching, which aims to provide an informative set of demonstrations. This
algorithm can partially reduce the computational cost of gathering
demonstrations and allows for an IRL algorithm that can learn more efficiently.

We are specifically interested in the application of IRL to discrete
optimization problems, a common problem to benchmark IRL algorithms. The
literature reports some applications of IRL in small grid-world problems
\citep{dong_towards_2024} and tabular problems, such as the Gymnasium
environments pendulum and cart-pole \citep{baimukashev_automated_2024}.
\citet{martinez-gil_using_2020} propose to compare the feature expectation
functions of the expert with those of the learner to evaluate the chosen reward
function. The selection of appropriate feature is very important in both IRL
and RL problems. In RL, reward function design has a large impact on the
behavior of the agent: a poorly chosen reward function may lead to unpredictable
and unwanted behavior \citep{francois-lavet_introduction_2018}. Similarly, the
success of IRL is closely related to the choice of features and accurate
descriptions of state transitions \citep{arora_survey_2021}.

\paragraph*{Contributions:}
In this paper, we consider a specific instance of the aforesaid data synthesis
problem that addresses minimum exposure to adverse environmental conditions
during travel to a fixed goal location. The environment is characterized by a
{threat field}, which is a time-varying strictly positive scalar field with
 higher intensity regions corresponding to potentially hazardous and
unfavorable conditions to the AV. We are interested in policies that minimize the AV's
exposure to threat. Note that the high-level description ``minimize exposure to
threat'' may be interpreted via several slightly different reward functions.
Indeed, different blackbox autonomy stacks for AVs that each claim to ``minimize
exposure to threat'' may in fact use slightly different cost functions that are
unknown to the user. We address the problem of synthesizing datasets of minimum
exposure paths that resemble a training dataset of paths. In this paper, we
focus on small datasets where we may quantify similarity of the synthesized
dataset to the training datasets by direct error calculations. However, the
proposed work is agnostic to the size of datasets and may be extended in the
future to larger training and synthetic datasets, where similarity is quantified
by statistical analysis.

The main contribution of this paper is an IRL model that solves the aforesaid
data synthesis problem. We consider time-invariant (static) as well as
time-varying (dynamic) threat fields. We find that the proposed IRL model
provides excellent performance for synthesizing paths from initial conditions
not seen in the training dataset when the threat field is the same as used for
training. Furthermore, we evaluate model performance on unseen threat fields and
find low error in that case as well. Finally, we demonstrate the model's ability
to synthesize distinct datasets when trained on different datasets with distinct
characteristics.

The rest of this paper is organized as follows. In \S\ref{sec-problem}, we
provide a precise problem formulation. In \S\ref{sec-method}, we describe the
details of the proposed IRL model and its training. In \S\ref{sec-results}, we
provide results of computational work regarding the performance of the proposed
model, and conclude the paper in \S\ref{sec-conclusions}.

\section{Problem Formulation}
\label{sec-problem}

Let $\real, \nat$ denote the sets of real and natural numbers, respectively. We
denote by $\intset{N}$ the set \{1, 2, \ldots, $N$\}, and by $\eye{N}$ the
identity matrix of size $N$, for any $N\in\nat.$ Finally, $\unif(a,b)$ denotes
the uniform distribution over the interval $\clint{a}{b},$ for $a, b \in \real$
and $a < b.$

Consider a closed square region in $\real[2]$ within which the AV operates,
overlaid with a grid consisting of $\nGridPts$ points. Time is discretized 
over $\nTimeSteps$ time steps.
The methods proposed in this work
do not necessitate uniform discretization, but for the sake of simplicity and clarity of
the main ideas, the numerical results are restricted to uniform discretizations
in space and time. 
The AV traverses grid points
according to the ``4-way adjacency rule,'' such that the adjacent points are
top, down, left, and right. We assume a constant speed such that the actor's
transitions to adjacent grid points occurs in a constant time step $\timeStep.$

The AV's motion is thereby modeled by a Markov Decision Process (MDP) $\langle
\mdpStateSet, \mdpActionSet, \mdpTransitionFcn \rangle.$ For static
environments, $\mdpStateSet \equiv \intset{\nGridPts}$ is a finite set of
observable states, with each state uniquely associated with the AV's location at
a grid point. For dynamic environments, $\mdpStateSet \equiv \intset{\nGridPts}
\times \nTimeSteps,$ and each state is uniquely associated with the AV's
location at a grid point at a unique time step. The grid point and/or time step
associated with a state $\mdpState$ are denoted $\xCoord[{\mdpState}]$ and
$\tCoord[{\mdpState}],$ respectively. $\mdpActionSet = \{\uparrow, \downarrow,
\leftarrow, \rightarrow\}$ is a set of possible actions that enables transition
to another state. $\mdpTransitionFcn: \mdpStateSet \times \mdpActionSet \times
\mdpStateSet \rightarrow \{0,1\}$ is a \mydef{transition function} such that
$\mdpTransitionFcn(\mdpState, \mdpAction, \mdpState')$ indicates whether or not
state $\mdpState'$ can be reached from $\mdpState$ under action $\mdpAction.$
This function is a restrictive form of the more general transition probability
distribution typically associated with an MDP.
For any $\mdpState \in \mdpStateSet,$ a \mydef{neighbor} state $\mdpState'$
is such that there exists an action $\mdpAction \in \mdpActionSet$ with
$\mdpTransitionFcn(\mdpState, \mdpAction, \mdpState') = 1.$

Transitions are characterized by a \mydef{reward function}
$\rewardFcn : \mdpStateSet \times \mdpAction \times \mdpStateSet
\to \real,$ which maps a state-action pair to a scalar reward.
A \emph{policy} $\policy$ maps the current state to the next state. It can be
either a deterministic function $\policy : \mdpStateSet \to \mdpActionSet$, or a
probability distribution over $\mdpActionSet.$ Each policy is associated with a
value function $\valueFcn:\mdpStateSet \to \real,$ which is the cumulative
expected reward for the policy $\policy$ for any initial state $\mdpState_0 \in
\mdpStateSet,$ namely:
\begin{align*}
	\label{eq:value}
	\valueFcn(\mdpState_0) &\defeq 
	\mathbb{E}_{\mdpState, \policy(\mdpState)} 
	\biggl[ \sum_{k=0}^\infty \discount^k
	\rewardFcn(\mdpState_k, \policy(\mdpState_k)) \biggl],
\end{align*}
where $\discount \in \lopint{0}{1}$ is a discount factor for future rewards.
For policy $\policy,$ every initial state $\mdpState_0$ is associated with a
\mydef{path}, namely a sequence of states $\gridPath = (\mdpState_0,
\mdpState_1, \ldots, \mdpState_k, \ldots )$ such that $\mdpState_k$ is
a neighbor of $\mdpState_{k-1}$ for each $k \in \nat.$ 
Because we consider deterministic transitions in the MDP,
every initial state is \emph{uniquely} associated with a path.
In this paper, we focus on paths of finite length that terminate
at a specific goal location, and therefore we consider $\discount = 1.$

The data synthesis problem of interest in this paper is described as follows.
Suppose we have a dataset $\trainingDataset = \{\gridPath^0, \gridPath^1,
\ldots, \gridPath^\nData\},$ of a finite number of paths traversed by an AV
under some policy. We would like to synthesize another dataset $\genDataset =
\{\gridPathGen^0, \gridPathGen^1, \ldots, \gridPathGen^\nGen\},$ that is in some
sense \emph{similar} to $\trainingDataset.$ Ideally, we want: (1) the size
$\nData$ of the original dataset $\trainingDataset$ to be ``small'' and (2) low
computational effort to synthesize $\genDataset$ of a ``large'' size $\nGen.$
Similarity of $\trainingDataset$ and $\genDataset$ may be evaluated in different
ways, including similarity of their statistical distributions or the closeness
of data points to a common benchmark.

We consider a specific instance of this general problem that addresses minimum
exposure to adverse environmental conditions during traversal to a fixed goal location
$\xCoordGoal.$ The environment is characterized
by a \mydef{threat field} $\threat : \real[2] \times \real_{\geq0}
\rightarrow\real_{>0}$, which is a time-varying strictly positive scalar field
 with higher intensity regions corresponding to potentially hazardous and
unfavorable conditions to the AV. We are interested in policies that minimize the AV's
exposure to threat. Note that the high-level description ``minimize exposure to
threat'' may be interpreted via several slightly different reward functions, as
we discuss later in this paper.

\section{Method}
\label{sec-method}

\begin{figure}
	\resizebox{\columnwidth}{!}{
		\begin{tikzpicture}[
			node distance=2cm and 3cm,
			every node/.style={draw, rectangle, 
				minimum size=1cm, text width = 3cm, text centered},
			->, >=stealth]
			
			\node (R) [thick, fill=lightgray, label=above:{\footnotesize Neural Network}]
			{Reward Function $\rewardFcn$};
			\node (samp) [below=1cm of R] {Sample Trajectories};
			\node (RL) [right=3cm of R] {Reinforcement Learning Algorithm};
			\node (pi) [thick, below right=1cm and 3cm of R, fill=lightgray, 
			label=below:{\footnotesize Neural Network}] {Optimal Policy $\policy$};
			\node (demos) [left=1.5cm of R] {Expert Demonstrations};
			
			\draw[->, thick] (R) to (RL);
			\draw[->, thick] (RL) to (pi);
			\draw[->, thick] (pi) to (samp);
			\draw[->, thick] (samp) to (R);
			\draw[->, thick] (demos) to (R);
			
		\end{tikzpicture}
	}
	\caption{Schematic illustration of IRL.}
	\label{fig-irl-schematic}
\end{figure}
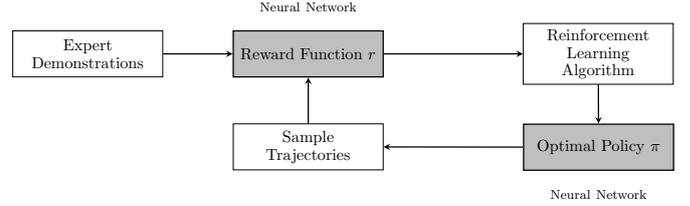

In this section, we describe the training of an IRL model using a training
dataset of paths. An IRL model simultaneously learns an optimal policy, ideally
$\policy^* = \arg\sup_\policy \valueFcn,$ and a reward function. To this end,
recall that a reinforcement learning (RL) model learns an optimal policy for a
\emph{given} reward function. One of the standard methods of training an RL
model is \emph{Q-learning.} The \emph{action-value function} $\QFcn :
\mdpStateSet \times \mdpActionSet \to \real$ maps a state-action pair to the
long-term expected reward with initial state $\mdpState_0$ and initial action
$\mdpAction_0$ under the policy $\policy.$ The optimal state-action value
function is $\optimalQFcn (\mdpState, \mdpAction) = \sup_\policy \QFcn
(\mdpState, \mdpAction),$ i.e., $\optimalQFcn (\mdpState, \mdpAction) \geq \QFcn
(\mdpState, \mdpAction)$ for each $\mdpState \in \mdpStateSet,$ each $\mdpAction
\in \mdpActionSet,$ and for any policy $\policy.$  It may be
shown~\citep{arora_survey_2021} that $\valueFcn[\policy^*](\mdpState) =
\sup_{\mdpAction \in \mdpActionSet} \optimalQFcn(\mdpState, \mdpAction).$

$\optimalQFcn$ must satisfy the Bellman equation, namely,
\begin{equation}
	\QFcn
	(\mdpState, \mdpAction) = \rewardFcn(\mdpState, \mdpAction, \mdpState') + 
	\discount \max_{\mdpAction \in \mdpActionSet}{\QFcn
		(\mdpState', \policy(\mdpState'))}.
	\label{eq:Bellman}
\end{equation}
RL attempts to learns $\optimalQFcn,$ after which the optimal policy
$\policy^*$ is defined as \citep{arora_survey_2021,kang_deep_2023}:
\begin{equation}
    \policy^*(\mdpState) = \arg\max_{\mdpAction \in \mdpActionSet} \optimalQFcn
    (\mdpState, \mdpAction).
    \label{eq:opt-policy}
\end{equation}

To train an RL model, the rewards for all state-action-state transitions
should be known.
However, designing an accurate reward function may be challenging in some
applications and/or an improperly designed reward function may lead to
convergence to a sub-optimal policy. When the reward is unavailable, it is
possible to train an IRL model that learns from data not only the optimal policy
(like RL) but also the reward function simultaneously, as illustrated in 
Fig.~\ref{fig-irl-schematic}. 
For an MDP
$\langle \mdpStateSet, \mdpActionSet, \mdpTransitionFcn \rangle$ and a dataset
$\trainingDataset$ described as above, IRL attempts to learn a reward function
$\rewardFcn$ and a policy $\policy^*$ that is optimal for the reward
$\rewardFcn,$ such that the errors between paths under $\policy^*$ 
and those in the dataset are minimized~\citep{martinez-gil_using_2020}.

Per~\citet{martinez-gil_using_2020}, we introduce a feature vector
$\featureFcn_\mdpState \in \real[2(\nAction + 1)]$ associated 
with each state that is used to formulate the reward function. 
Specifically:
\begin{align*}
	\featureFcn_\mdpState &\defeq \left(
	\threat( \xCoord[\mdpState], \tCoord[\mdpState] ),
	\| \xCoord[\mdpState] - \xCoordGoal \|, 
	\threat( \xCoord[\mdpState'], \tCoord[\mdpState'] ),
	\| \xCoord[\mdpState'] - \xCoordGoal \|, \ldots  \right),
\end{align*}
where $\mdpState'$ indicates each neighbor of $\mdpState.$
We assume that the reward function to be learned is a linear combination
of some of the features, namely:
\begin{align}
	\label{reward}
	\rewardFcn(\mdpState, \mdpAction, \mdpState') &\defeq 
	\wtReward_1 \threat( \xCoord[\mdpState'], \tCoord[\mdpState'] ) + 
	\wtReward_2 \| \xCoord[\mdpState'] - \xCoordGoal \|.
\end{align}
In other words, we model the reward function as a linear combination of
the threat intensity and distance from the next state to the goal location.

%

The training process of the proposed IRL model is described in
Algorithm~\ref{alg:IRL}, which provides the reward learning algorithm, and
Algorithm~\ref{alg:Q-Learning}, which describes the Q-learning algorithm.
The iterative interactions of the two algorithms are illustrated in
Fig.~\ref{fig:irl_process}.

\tikzstyle{process} = [rectangle, rounded corners, minimum width=1cm, minimum height=0.75cm, text 
centered, draw=black, fill=white, thick]
\tikzstyle{arrow} = [line width=1mm,->,>=stealth]

\subsection{IRL and Reward Learning}

The proposed IRL training algorithm iteratively updates the reward weights
$(\wtReward_1, \wtReward_2)$ based on errors between a sample set of optimal
paths synthesized using the learned reward and the paths in the training dataset
$\trainingDataset$ of size~$\nData.$ To this end, we first calculate a feature
expectation vector $\trainingFeatEx \in \real[2]$ over the training data as
follows. For each finite path $\gridPath^n = (\mdpState_0^n, \ldots,
\mdpState_k^n, \ldots, \mdpState_{\gridPathLength[n]}) \in \trainingDataset,$ we
calculate an aggregate feature $\aggregateFeature_n \defeq
\sum_{k=0}^{\gridPathLength[n]} \featureFcn_{\mdpState_k^n}[1:2].$ The feature
expectation vector is the average of $\aggregateFeature_n$ over the dataset:
\begin{align*}
	\trainingFeatEx &\defeq 
	\frac{1}{\nData} \sum_{n = 1}^{\nData} 
	\aggregateFeature_n = 
	\frac{1}{\nData} \sum_{n = 1}^{\nData} 
	\sum_{k=0}^{\gridPathLength[n]} \featureFcn_{\mdpState_k^n}[1:2].
\end{align*}

At the $i\msup{th}$ training iteration, Algorithm~\ref{alg:IRL} executes
Q-learning (Algorithm~\ref{alg:Q-Learning}) for $\QLMaxEpisodes$ episodes
to find an optimal policy
$\policy_{\wtQFcn^i}$ using the reward function at the current iteration.
Here,~$\wtQFcn^i$ denotes the parameters of the Q-learning model at
iteration~$i,$ discussed in \S\ref{ssec-qlearning}. Using this optimal
policy~$\policy_{\wtQFcn^i}$ and a sample of initial states taken
from~$\mdpStateSet,$ we find a set of $\nGen$ paths resulting from this policy.
Each of these sample paths either terminates at the goal location or after a
prespecified maximum number of transitions~ $\IRLMaxPathIter \in \nat.$ The
magnitude of error between $\trainingFeatEx$ and the feature expectation vector
over these paths provides an update to the reward weights~$\wtReward^i$ and the
iterations continue.

Algorithm~\ref{alg:IRL} is initialized with an initial guess for the reward
weights, e.g., $\wtReward^0 \defeq \trainingFeatEx.$ The
parameters~$\wtQFcn^0$ are initialized to arbitrary values. The algorithm
terminates when the error reduces below a desired threshold $\IRLErrorThreshold
> 0$, or returns failure to do so after a prespecified maximum number of
iterations~$\IRLMaxIterations \in \nat.$ The hyperparameter~$\IRLLearningRate$
in Algorithm~\ref{alg:IRL} is the learning rate. The numerical values of
$\wtReward^i$ and $\featEx[i]$ are scaled, shifted, and clipped appropriately to
ensure convergence of the deep Q-learning algorithm.

\begin{algorithm}
\caption{Iterative schema for IRL}
\label{alg:IRL}
	\begin{algorithmic}
	\State $i \defeq 0$;
	\State Set $\featEx[0] \defeq \wtReward^0$ and $\featExTmp \defeq (0,0)$;
	\While{$\| \featExTmp - \trainingFeatEx \| \geq \IRLErrorThreshold$ 
		and $i \leq \IRLMaxIterations$}
	    \For {$\QLMaxEpisodes$ episodes}
	    	\State $\policy_{\wtQFcn^i} \defeq $ output of Deep Q-learning 
	    	(Algorithm~\ref{alg:Q-Learning})
	    \EndFor
	    \For {$n = 1, 2, \ldots, \nGen$}
	    	\State Sample initial state $\mdpState^n_0$ from $\mdpStateSet$;
	    	\State $k \defeq 1;$ $\aggregateFeature_n \defeq (0, 0)$;
	    	\While {$k < \IRLMaxPathIter$}
	    		\State Find neighbor $\mdpState^n_k$ such that 
	    		$$\mdpTransitionFcn(\mdpState^n_{k-1}, 
	    		\policy_{\wtQFcn^i}(\mdpState^n_{k-1}), 
	    		\mdpState^n_k) = 1;$$
				\State $\aggregateFeature_n \defeq \aggregateFeature_n + 
				\featureFcn_{\mdpState_k^n}[1:2]$;
	    		\If {$\xCoord[\mdpState^n_k] = \xCoordGoal$}
	    			\State break;
	   	 	\EndIf
	   	 	\State $k \defeq k + 1$;
	    	\EndWhile
	    \EndFor
	    \State Calculate $\featExTmp \defeq \frac{1}{\nGen} \sum_{n=1}^{\nGen} 
	    \aggregateFeature_n$ and $\featError \defeq \featExTmp - \featEx[i]$;
	    \State ~ 
	    \State $\featEx[i+1] \defeq \featEx[i] + 
	    \frac{1}{\| \featError \| } \innerp{\featError}{\featEx[i]}  \featError$;
	    \State ~ 
	    %
	    \State Update $\wtReward^{i+1} \defeq \IRLLearningRate 
	    \featEx[i+1] + (1 - \IRLLearningRate) \wtReward^i$;
	    \State $i \defeq i + 1$;
	\EndWhile
	\State 
	\Return $\wtReward \defeq \wtReward^i$ and $\policy^* \defeq \policy_{\wtQFcn^i}$
	\end{algorithmic}
\end{algorithm}

\subsection{Deep Q-Learning}
\label{ssec-qlearning}

\begin{figure}
	\centering
	\resizebox{\columnwidth}{!}{	
		\begin{tikzpicture}[scale=0.6]
			\draw[arrow] (0, 0) arc[start angle=85, end angle=-240, x radius=6cm, y radius=3cm];
			\node[process] (A) at (-0.5, -0.5) {\shortstack{DQL: learn policy $\policy_\wtQFcn$}};
			\node[process] (B) at (5, -3) {\shortstack{Synthesize sample paths \\ and calculate \\
					feature expectation $\featEx$}};
			\node[process] (D) at (-0.5, -5.5) {\shortstack{Calculate loss $\featError$}};
			\node[process] (E) at (-6,-3) {\shortstack{IRL: Update \\ reward weights $\wtReward$}};
		\end{tikzpicture}
	}
	
	\caption{Illustration of the IRL and DQL training processes.}
	\label{fig:irl_process}
\end{figure}
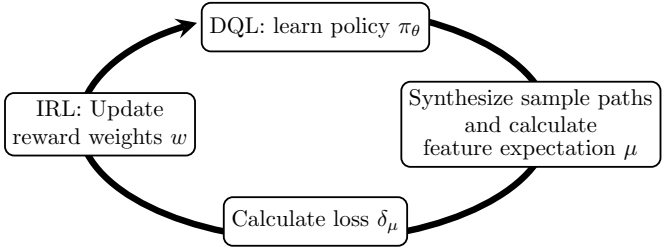

We implement a Q-learning model as a multi-layer perceptron neural network,
henceforth called the deep Q-learning (DQL). The parameters (weights and biases)
in this DQL model are collectively denoted~$\wtQFcn.$ The proposed DQL training
process, described in Algorithm~\ref{alg:Q-Learning}, is a modified form of the
work by~\citet{mnih_playing_2013}.

Algorithm~\ref{alg:Q-Learning} maintains parameters of the DQL model of the
state-action value function, denoted $Q,$ as well as a ``mirror'' model $Q'.$ Whereas
$Q$ is the desired DQL model, maintaining $Q'$ helps in the convergence of the
training of~$Q.$ The parameters of $Q'$ are denoted~$\wtQFcnP.$
Note that the parameters of $Q$ and $Q'$ are initialized only
once in Algorithm~\ref{alg:IRL} and then maintained in memory between successive
calls to Algorithm~\ref{alg:Q-Learning}. Because the $Q$ function depends on the
reward function, this DQL model is parametrized by the reward weights~$\wtReward^i.$
To explicitly indicate the parameters, we denote the two DQL models by
$Q_{\wtQFcn}$ and $Q'_{\wtQFcnP},$ respectively.

The parameters of $Q'_{\wtQFcnP}$ are updated either using a soft update
or a hard reset, whereas the the weights based on~$Q_{\wtQFcn}.$ 
At its $j\msup{th}$ training iteration, we select an action either
randomly or by maximizing the currently learned~$Q_{\wtQFcn}.$ This choice is based on a
decreasing probabilistic threshold $\selectionProb$ of selecting a random
action~\citep{mnih_playing_2013}, which similar to the random exploration step
in evolutionary optimization algorithms like simulated annealing.
A soft (gradient descent-like) update of $\wtQFcnP$ occurs at each iteration,
whereas a hard reset occurs whenever a mean-squared error loss reduces below
a prespecified threshold~$\lossThreshold.$

Whereas the $Q_{\wtQFcn}$ and $Q'_{\wtQFcnP}$ functions map state-action pairs to a value,
it is numerically convenient set up the DQL models to accept the feature
vector $\featureFcn_\mdpState$ as input instead of an $(\mdpState, \mdpAction)$ pair.
Note that the same input information is preserved in either case.

Algorithm~\ref{alg:Q-Learning} executes for a
fixed number of iterations~$\QLMaxIterations.$ 

\begin{algorithm}
\caption{Deep Q-Learning at $i\msup{th}$ IRL iteration}
\label{alg:Q-Learning}
	\begin{algorithmic}
	\State $j \defeq 0$
	\While {$j \leq \QLMaxIterations$}
		\State $\rewardPlusVec \equiv \emptyset$
		\State	Sample $z \sim \unif(0,1)$
		\For {$\mdpState \in \mdpStateSet,$}
			\If {$z < \selectionProb$}
			  	\State Randomly select $\mdpAction^*_\mdpState \in	\mdpActionSet$;
			\Else {}
			  	\State Find
			  	$\mdpAction^*_\mdpState \defeq \arg \max_{\mdpAction \in \mdpActionSet} 
			  	Q_{\wtQFcn}(\mdpState, \mdpAction; \wtReward^i);$
			\EndIf
			\State Find $\mdpState'$ such that
			$\mdpTransitionFcn(\mdpState, \mdpAction^*_\mdpState, \mdpState') = 1;$
			\State Calculate the loss $$\rewardPlus_\mdpState \defeq \rewardFcn(\mdpState, 
			\mdpAction^*_\mdpState, \mdpState') - Q_{\wtQFcn}(\mdpState, 
			\mdpAction^*_\mdpState; \wtReward^i) + \discount \max_{\mdpAction^*_{\mdpState'} 
				\in \mdpActionSet} Q'_{\wtQFcnP}(\mdpState', \mdpAction^*_{\mdpState'}; 
				\wtReward^i);$$
			\State Append $\rewardPlus_\mdpState$ to $\rewardPlusVec$
		\EndFor
		\State Update $\wtQFcn \defeq \arg\min_{\wtQFcn} \| \rewardPlusVec \|^2$
		\State Soft update: $\wtQFcnP \defeq \DQLLearningRateP \theta_{q} + 
		(1 - \DQLLearningRateP) \wtQFcnP;$
		\State $j \gets j+1$;
	\EndWhile
	\If {$\| \rewardPlusVec \| \leq \lossThreshold$}
		\State Hard reset: $\wtQFcnP \defeq \wtQFcn$
	\EndIf 
	\State \Return $\policy_{\wtQFcn^i}(\mdpState) \defeq  
	\arg\max_{\mdpAction \in \mdpActionSet} Q_{\wtQFcn}(\mdpState, \mdpAction; \wtReward^i).$
	\end{algorithmic}
\end{algorithm}

At the $j\msup{th}$ iteration, the selection probabilistic threshold
$\selectionProb$ is chosen as
\begin{equation*}\label{eq:epsilon}
	\selectionProb \defeq \selectionProbA + (\selectionProbB - 
	\selectionProbA) \exp(-\myfracB{j}{\decayRate}),
\end{equation*}
where $\selectionProbA$, $\selectionProbB$, and $\decayRate$ are  
hyperparameters to be tuned.



\subsection{Tuning Hyperparameters}
\label{ssec-hyperparameters}

To tune the relevant hyperparameters, it is important to consider the stability
of training each of the neural network models mentioned above.

In Algorithm~\ref{alg:IRL}, the error threshold $\IRLErrorThreshold$ dictates
the level of accuracy for the reward function loss. In other words, it
determines when the policy is sufficiently trained. The maximum number of
transitions allowed in a path, $\IRLMaxPathIter,$ determines the length of the
feature function for each path. A larger value of $\IRLMaxPathIter$ penalizes
non-convergent paths more than a smaller value. Ideally, the number of points
calculated according to the learned policy should be significantly larger than
the maximum ideal path length. This appropriately penalizes paths that remain
within the field yet do not converge and encourages the algorithm to find
convergent solutions.

In Algorithm~\ref{alg:Q-Learning}, the learning rates $\DQLLearningRateP$
and $\DQLLearningRate$ used to
update $\wtQFcnP$ and $\wtQFcn,$ respectively,
should be sufficiently small that the network is able to
converge. The selection threshold hyperparameters $\selectionProbB$,
$\selectionProbA$, and $\decayRate$ determine the length of random exploration
to the time spent refining the current policy.

The values of various hyperparameters, tuned after several numerical experiments,
are provided in Table~\ref{tab:tuning}.

\begin{table}[htbp]
\caption{Tuned Values of Hyperparameters}
\centering
\begin{tabular}{ll|ll|ll}
	\toprule
	\textbf{Hyperp.} & \textbf{Value} & \textbf{Hyperp.} & 
	\textbf{Value} & \textbf{Hyperp.} & \textbf{Value} \\
	\midrule
	$\IRLMaxPathIter$ & 500 & $\QLMaxEpisodes$ & 300 & $\QLMaxIterations$ & 25 \\
	%
	$\IRLLearningRate$ & 0.01 & $\DQLLearningRate$ & 0.005 & $\DQLLearningRateP$ & 0.001 \\
	$\selectionProbA$ & 0.051 & $\selectionProbA$ & 0.95 & $\decayRate$ &  500 \\
	$\IRLErrorThreshold$ & 0.01 & \\
	\bottomrule
\end{tabular}
\label{tab:tuning}
\end{table}

\section{Results}
\label{sec-results}

\begin{figure}
	\centering
	\includegraphics[width=0.7\columnwidth]{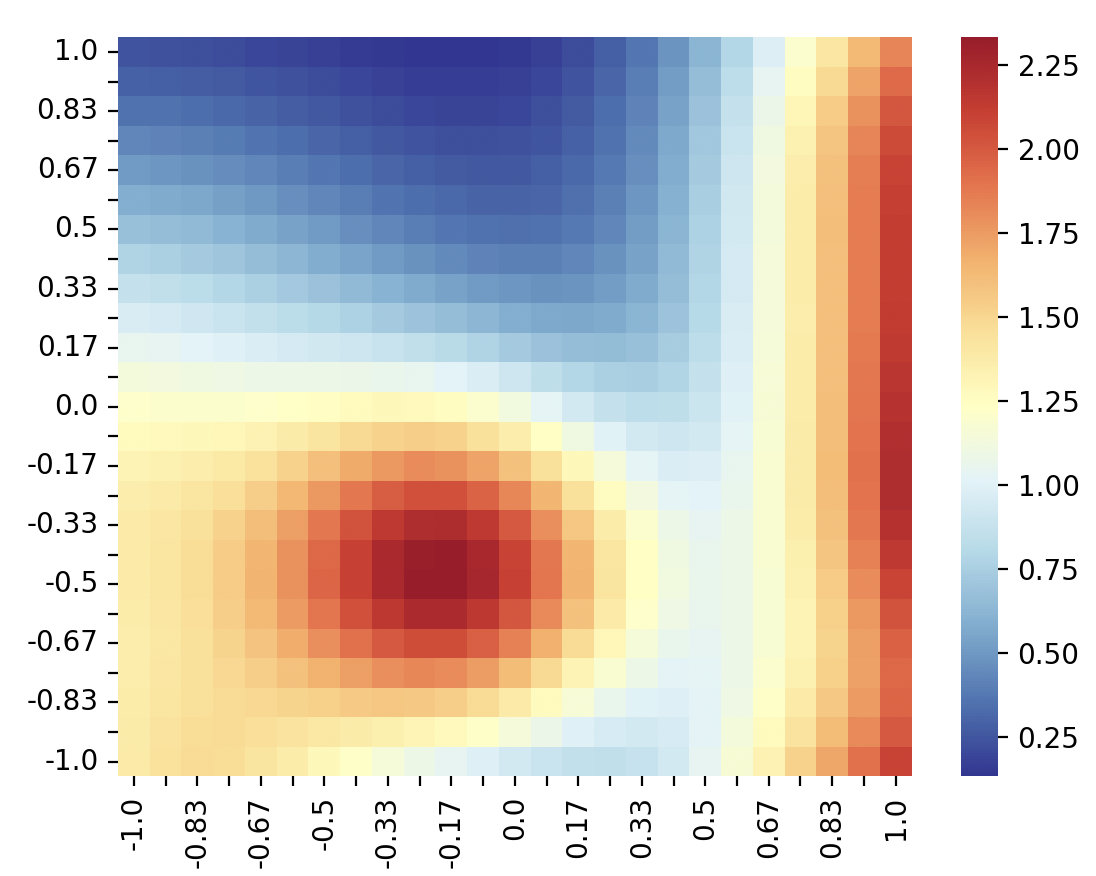}
	\caption{Sample static threat field.}
	\label{fig:blank-threat}
\end{figure}

\begin{figure}
	\centering
	\includegraphics[width=0.7\columnwidth]{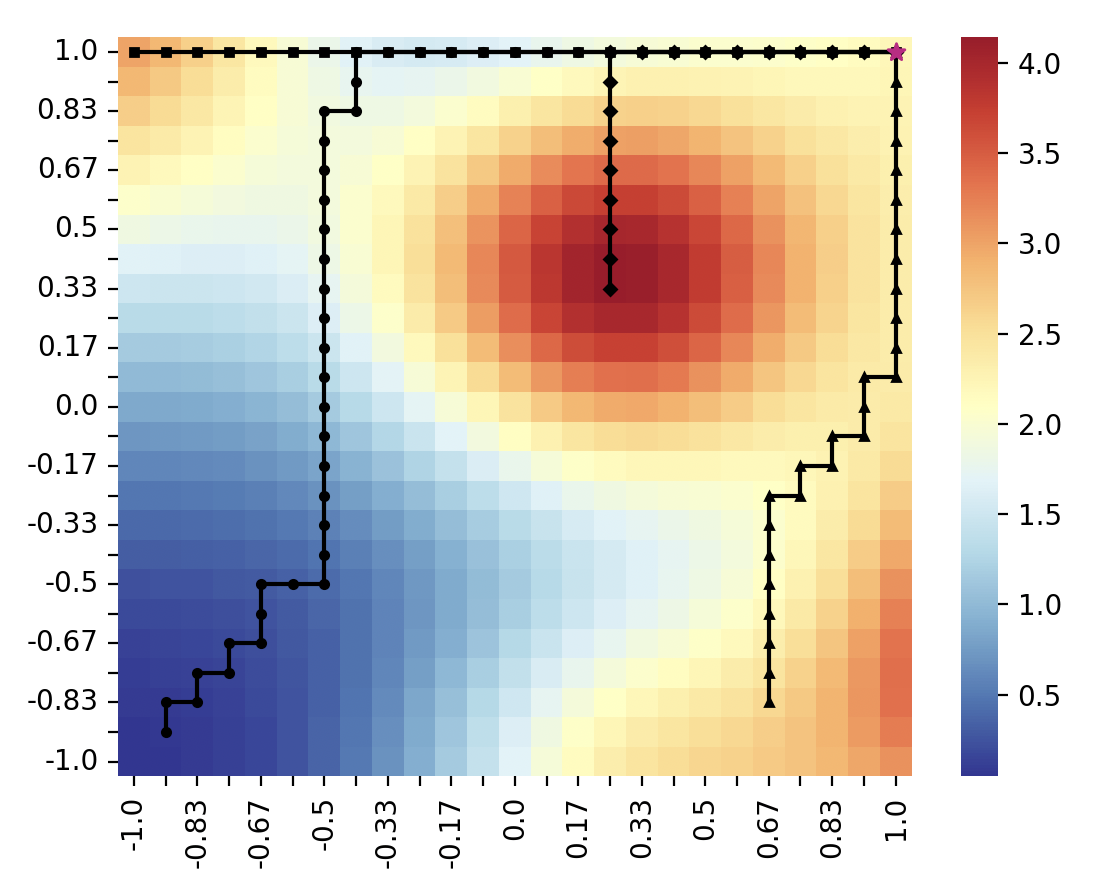}
	\caption{Examples of minimum threat exposure paths in the training 
		dataset~$\trainingDataset$ for the static case.}
	\label{fig:expert-paths}
\end{figure}

\begin{figure*}[h]
	\centering
	\includegraphics[width=0.9\textwidth]{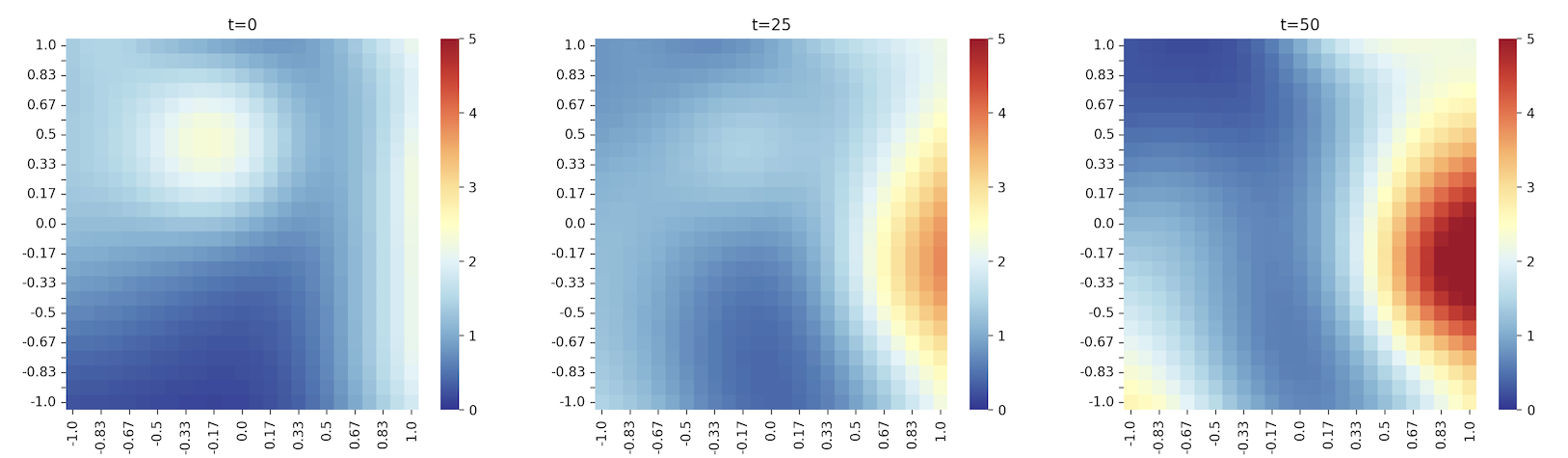}
	\caption{Sample evolution of a dynamic threat field.}
	\label{fig:threat_evolution}
\end{figure*}

We perform numerical experiments with static and dynamic threat fields, as
illustrated in Figs.~\ref{fig:blank-threat} and~\ref{fig:threat_evolution}. The
spatial discretization is over $\nGridPts = 625$ points in a uniform $25 \times
25$ grid, whereas temporal discretiztion is over $\nTimeSteps = 50$ time steps.
Various examples of static fields are constructed using a finite series of
radial basis functions with randomly chosen coefficients. To construct evolving
threat fields, as illustrated in Fig.~\ref{fig:threat_evolution}, we use linear
interpolation between two randomly generated static fields. The environment is
restricted to a non-dimensional square region $\clint{-1}{1} \times
\clint{-1}{1}.$ The goal location is fixed at $\xCoordGoal = (1,1).$

First, we study the performance of the proposed IRL model for ``pure'' minimum
threat exposure paths. This means, in the static case, each path $\gridPath =
(\mdpState_0, \mdpState_1, \ldots, \mdpState_{\gridPathLength})$ in the dataset
$\trainingDataset$ satisfies $\xCoord[{\mdpState_{\gridPathLength}}] =
\xCoordGoal$ and minimizes  $\sum_{k = 0}^{\gridPathLength}
\threat(\xCoord[{\mdpState_k}])$ over all paths between the same start and goal
locations. Figure~\ref{fig:expert-paths} illustrates some such paths. Minimum
threat exposure is similarly defined for dynamic threat fields, i.e., an expert
path minimizes $\sum_{k = 0}^{\gridPathLength} \threat(\xCoord[{\mdpState_k}],
\tCoord[{\mdpState_k}]),$ where $\tCoord[{\mdpState_{k}}] =
\tCoord[{\mdpState_{k-1}}] + \timeStep.$ The training dataset is assumed to
be noiseless, i.e., the paths in $\trainingDataset$ are exactly optimal.

In each case, we validate our IRL algorithm by comparing the costs of paths
generated by the trained policy $\policy_{\wtQFcn}$ against true optimal paths
for various initial states. If a true optimal path for a particular initial
state is not already provided in $\trainingDataset,$ we can easily calculate it
using Dijkstra's algorithm.

The software developed for training the models reported in this paper is available: \url{https://github.com/aeballentine/IRL-minimum-threat-navigation.git}.

\subsection{Static Threat Fields}
\label{ssec-static-results}

The initial guess for the reward function weights is $\wtReward^0 = (-1,-1),$
i.e., $\rewardFcn(\mdpState, \mdpAction, \mdpState') = - \threat(
\xCoord[\mdpState'], \tCoord[\mdpState'] ) - \| \xCoord[\mdpState'] -
\xCoordGoal \|,$ which is an intuitively reasonable guess for the reward for
minimum threat exposure.

\begin{figure}[ht]
	\centering
	\begin{subfigure}{0.49\columnwidth}
		\centering
		\includegraphics[width = \textwidth]{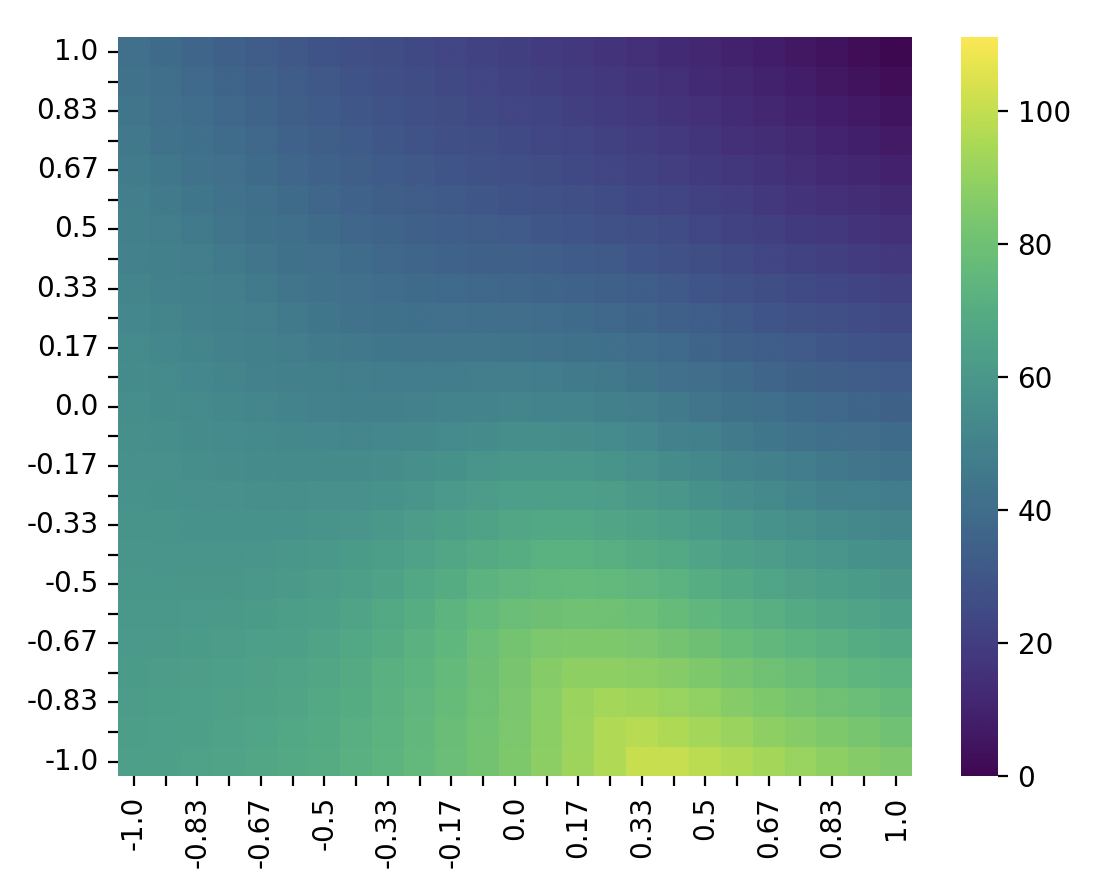}
		\caption{Value function found by Dijkstra's algorithm.}
		\label{fig:IRL_dijkstra}
	\end{subfigure}
	\begin{subfigure}{0.49\columnwidth}
		\centering
		\includegraphics[width = \textwidth]{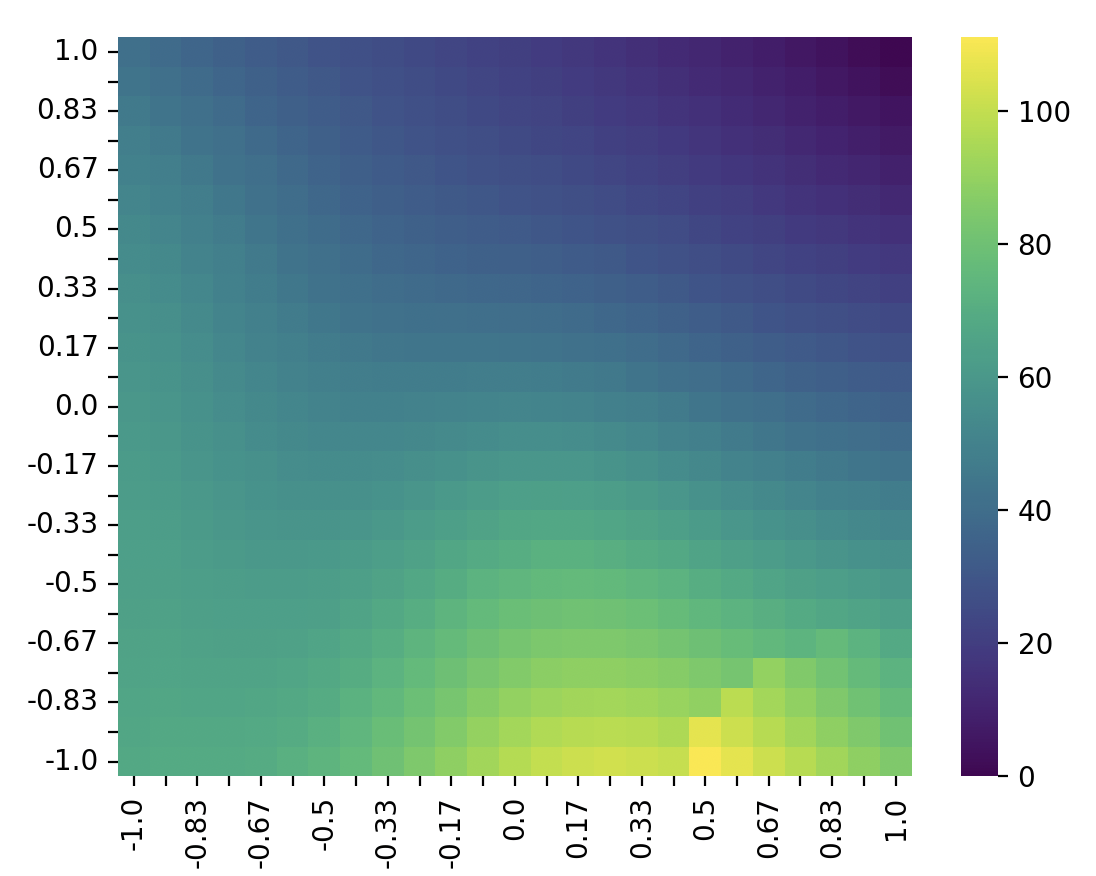}
		\caption{Value function due to the learned policy.}
		\label{fig:IRL_nn}
	\end{subfigure}
	\begin{subfigure}{0.7\columnwidth}
		\centering
		\includegraphics[width = \textwidth]{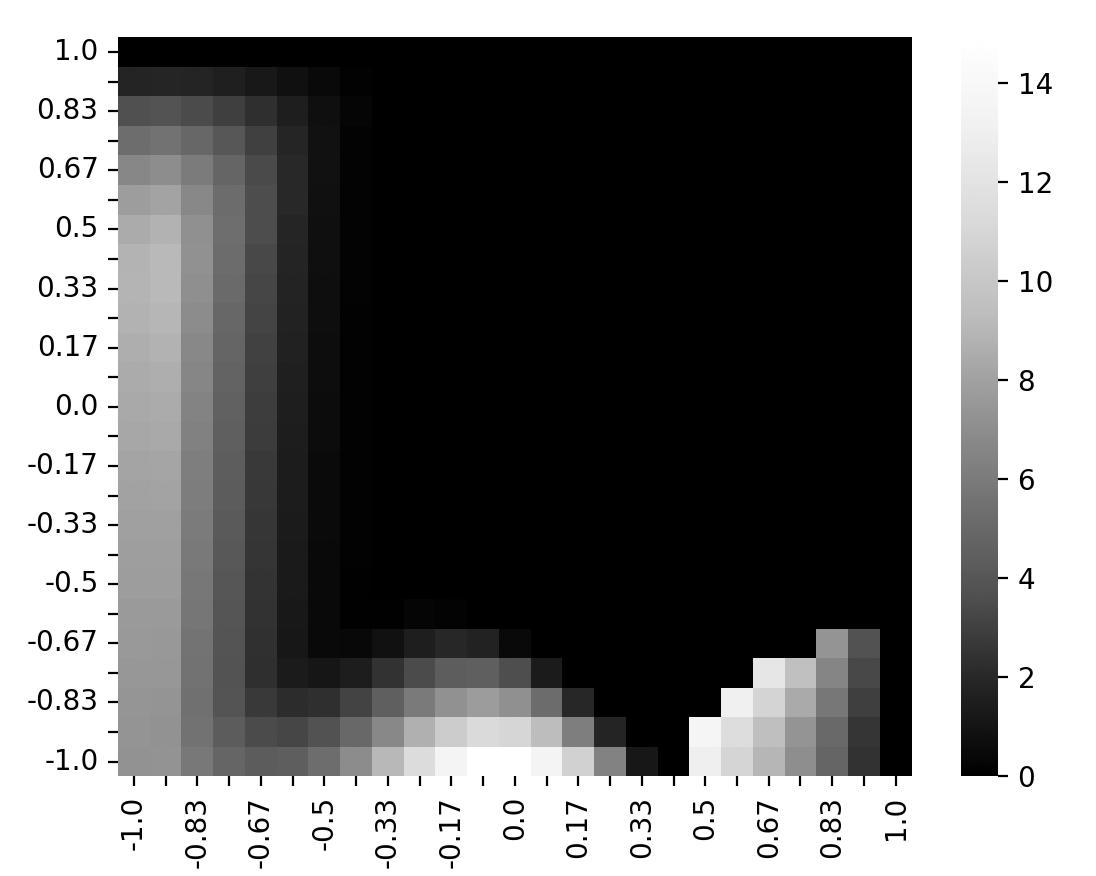}
		\caption{Percent error between the two value functions.}
		\label{fig:IRL_error}
	\end{subfigure}
	\caption{Performance of the proposed IRL model for a static environment.}
	\label{fig:combined-IRL}
\end{figure}

Figure~\ref{fig:combined-IRL} provides a comparsion of the value functions,
i.e., costs to reach the goal $\xCoordGoal$ learned by the proposed IRL model
(Fig.~\ref{fig:IRL_nn}) and as calculated using Dijkstra's algorithm
(Fig.~\ref{fig:IRL_dijkstra}). Figure~\ref{fig:IRL_error} shows the error
between the two value functions. Note that the error is in general small, and
the maximum error is around~15\% for initial states farther away from the goal
location, which indicates a good performance of the proposed IRL model.

\begin{figure}[h]
	\centering
	\begin{subfigure}{0.49\columnwidth}
		\centering
		\includegraphics[width=\textwidth]{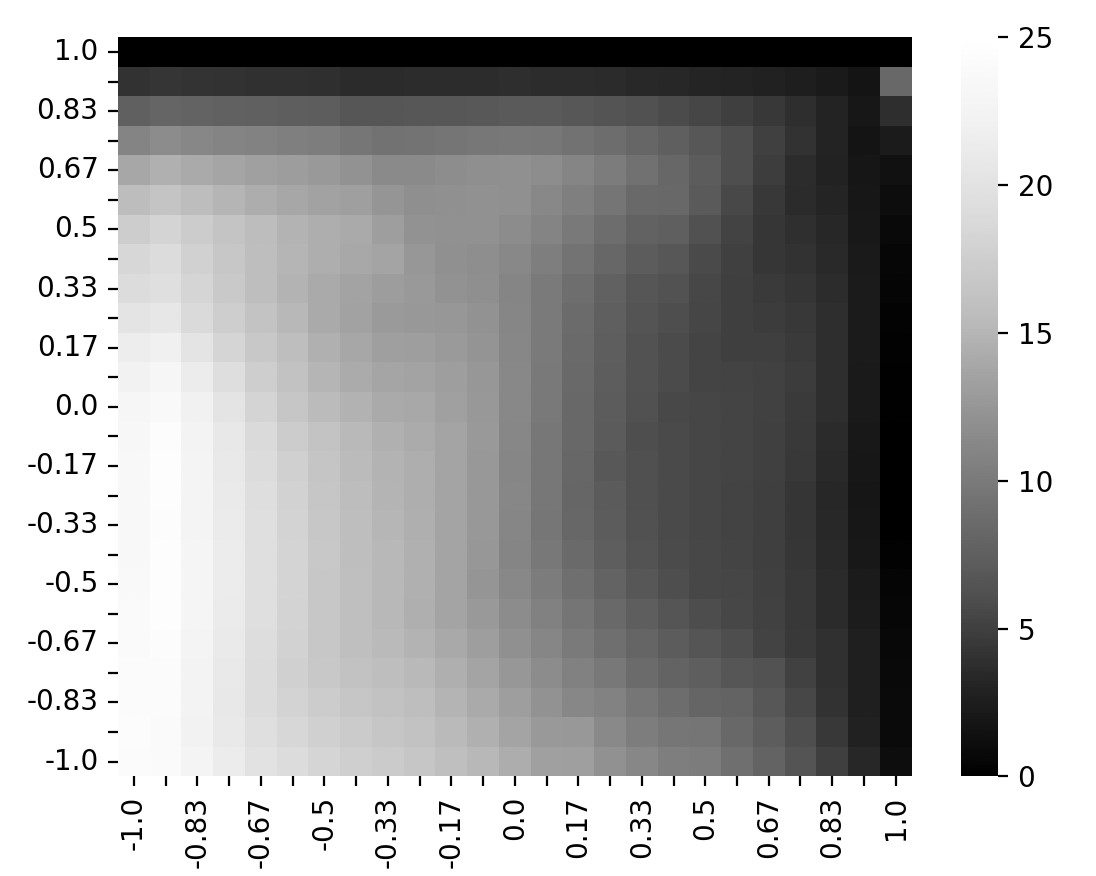}
		\caption{Mean error.}
		\label{fig:percent_error_static}
	\end{subfigure}
	\begin{subfigure}{0.49\columnwidth}
		\centering
		\includegraphics[width=\textwidth]{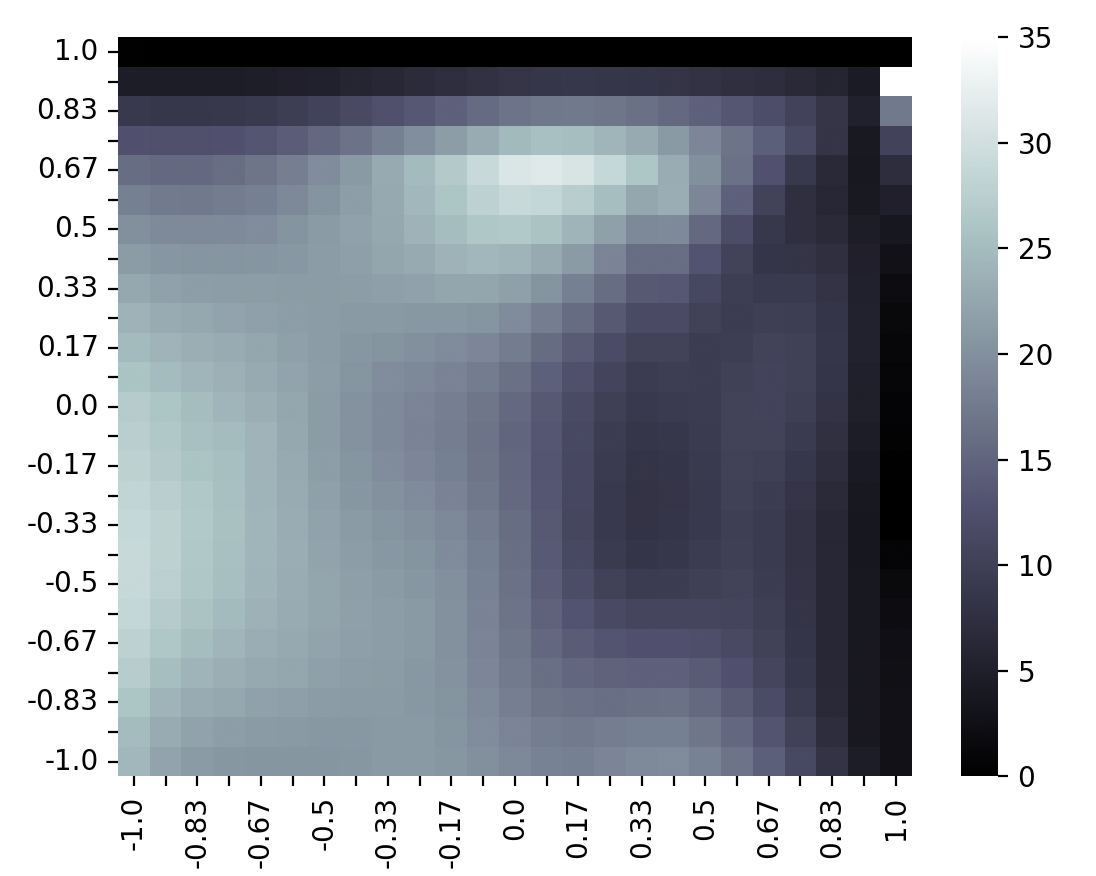}
		\caption{Error standard deviation.}
		\label{fig:sd_static}
	\end{subfigure}
	\caption{Performance over 22 training instances: mean and 
		standard deviation of percent error for the static case.}
	\label{results:static}
\end{figure}

Figure~\ref{results:static} gives the mean and percent error over multiple
training instances, which are trained on distinct threat fields. 
Recall that the probabilistic threshold $\selectionProb$
introduces inherent stochasticity in the DQL training, which may lead to
different policies. Note that some policy neural networks are disregarded if the
algorithm is not able to converge to the finish from all possible starting
locations in the threat field. We chose to filter out these results as they
result from sub-optimal training of the neural network during the Q-learning
phase. Figure~\ref{fig:percent_error_static} shows that the father away from the
goal location, the worse the policy performs in general. For starting locations
which are along the boundary of the graph, the IRL model is easily able to mimic
the optimal path, leading to low error along the top and right borders of the
threat field.

\subsection{Time-Varying Threat Fields}

\begin{figure}
	\centering
	\begin{subfigure}{0.49\columnwidth}
		\centering
		\includegraphics[width=\textwidth]{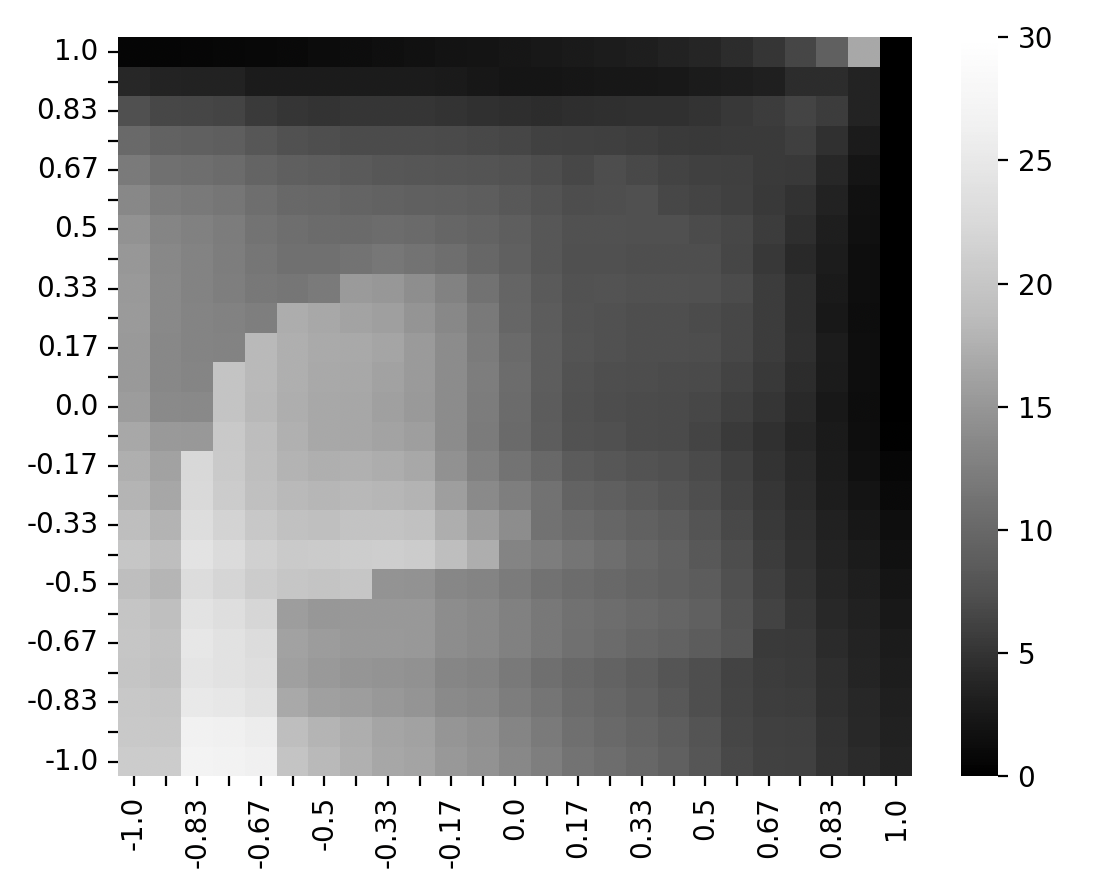}
		\caption{Mean error.}
		\label{fig:percent_error_varying}
	\end{subfigure}
	\begin{subfigure}{0.49\columnwidth}
		\centering
		\includegraphics[width=\textwidth]{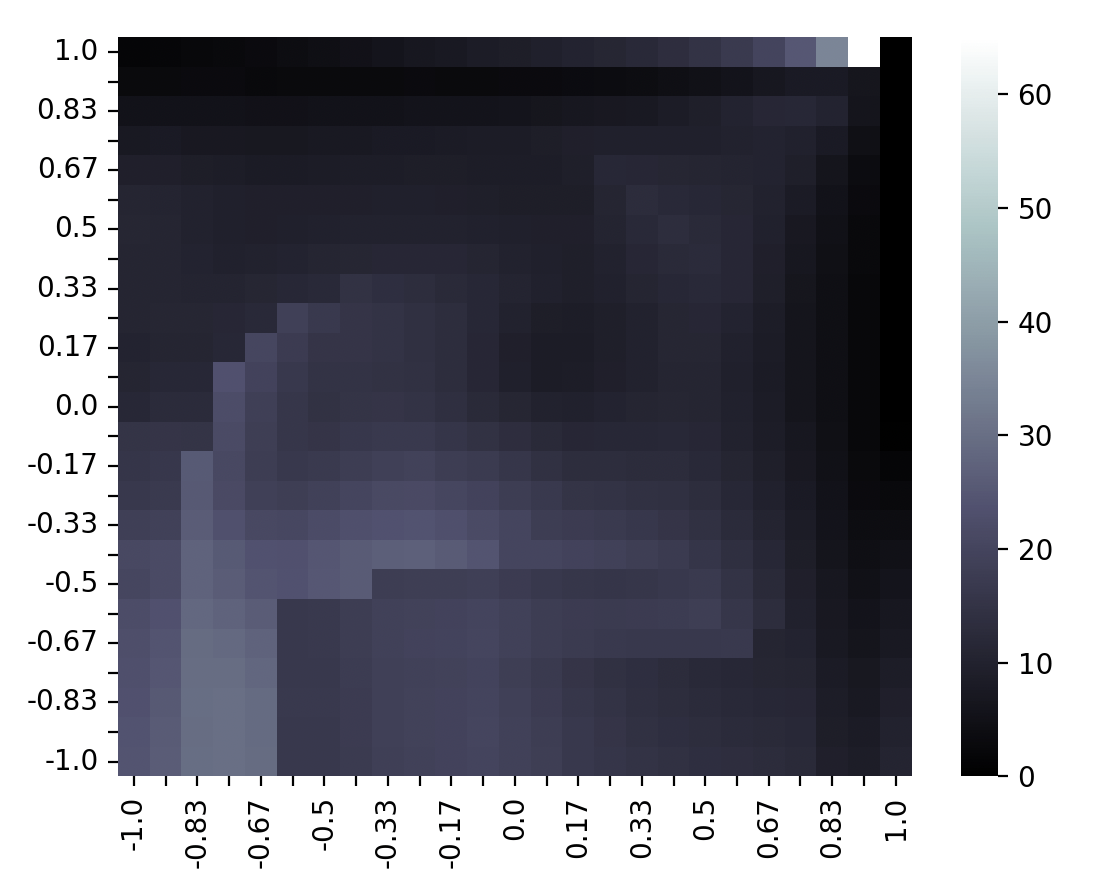}
		\caption{Error standard deviation.}
		\label{fig:sd_varying}\end{subfigure}
	\caption{Performance over 15 training instances: 
		mean and std. dev. of percent error for the 
		dynamic case.}
	\label{fig:results_varying}
\end{figure}

Learning to navigate through time-varying threat fields involves the same
training techniques, but requires a new definition of the transitions between
locations, as movements process an agent through the threat field as well as
through time. We find that the inclusion of time-varying threat fields does not
change the performance of the proposed IRL model.
Figure~\ref{fig:results_varying} illustrates the performance of 15 trained
policy networks by comparing the cost from each starting location to that of the
optimal path.

Similar to the static case, the algorithm performs well closer to the finish
point, but does not perform as well at locations which are father away. Also
similarly, the algorithm performs better, in general, along the upper and right
edge, as the path to the finish is relatively straightforward and less dependent
on the maximum intensity threat through the middle of the field, where there are
two feasible actions that lead closer toward the goal.

\subsection{Generalization to New Environments}

\begin{figure}
	\centering
	\begin{subfigure}{0.49\columnwidth}
		\centering
	        \includegraphics[width=\textwidth]{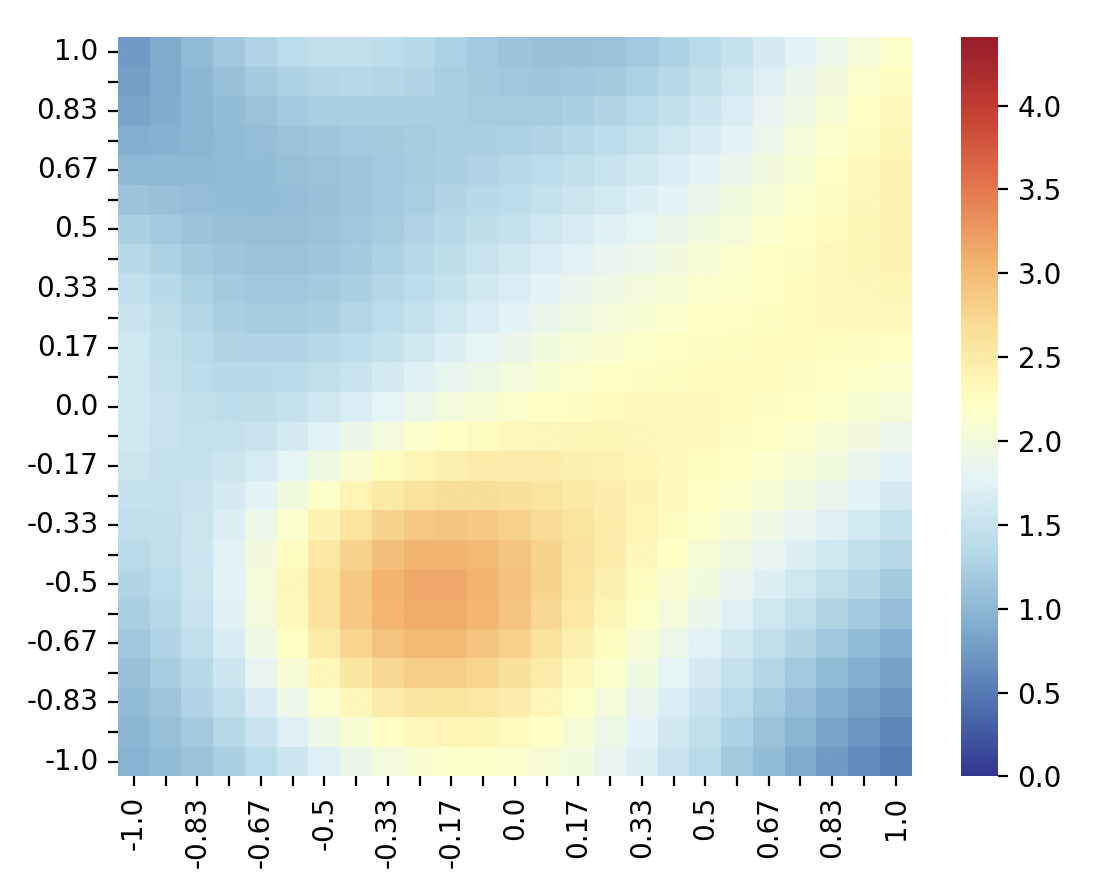}
	        \caption{Threat field for training.}
	        \label{fig:training_threat}
	\end{subfigure}
	\begin{subfigure}{0.49\columnwidth}
		\centering
	        \includegraphics[width=\textwidth]{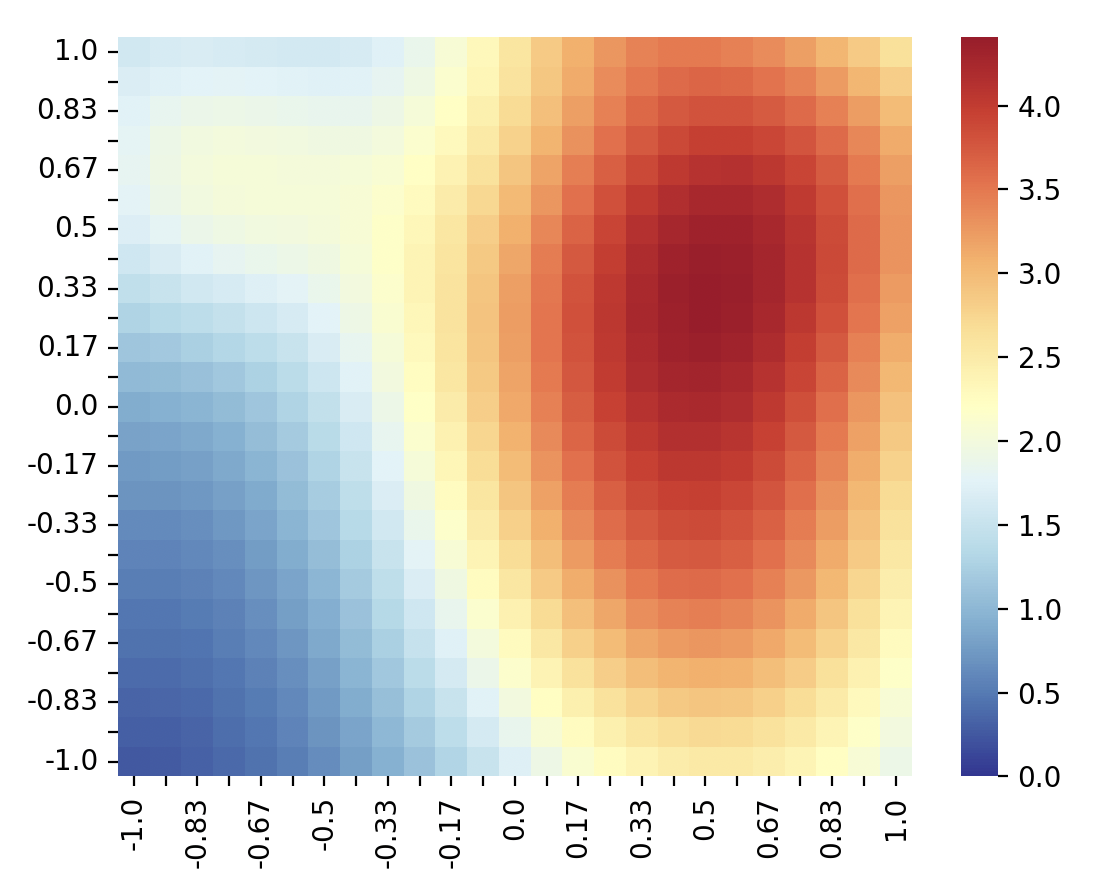}
	        \caption{Threat field for testing.}
	        \label{fig:testing_threat}
	\end{subfigure}
	\caption{Different threat fields training and testing.}
	\label{fig:generalization-threat}
\end{figure}

The results in the previous two subsections consider the overall error only for
one threat field, i.e., the one provided during training. Here, we consider two
evaluations: that of the threat field provided during training and a new threat
field that was not seen during training. As Figs.~\ref{fig:training_threat} and
~\ref{fig:testing_threat} illustrate, threat fields -- and, consequently,
optimal policies -- can differ significantly.

\begin{figure}
	\begin{subfigure}{0.49\columnwidth}
		\centering
	        \includegraphics[width=\textwidth]{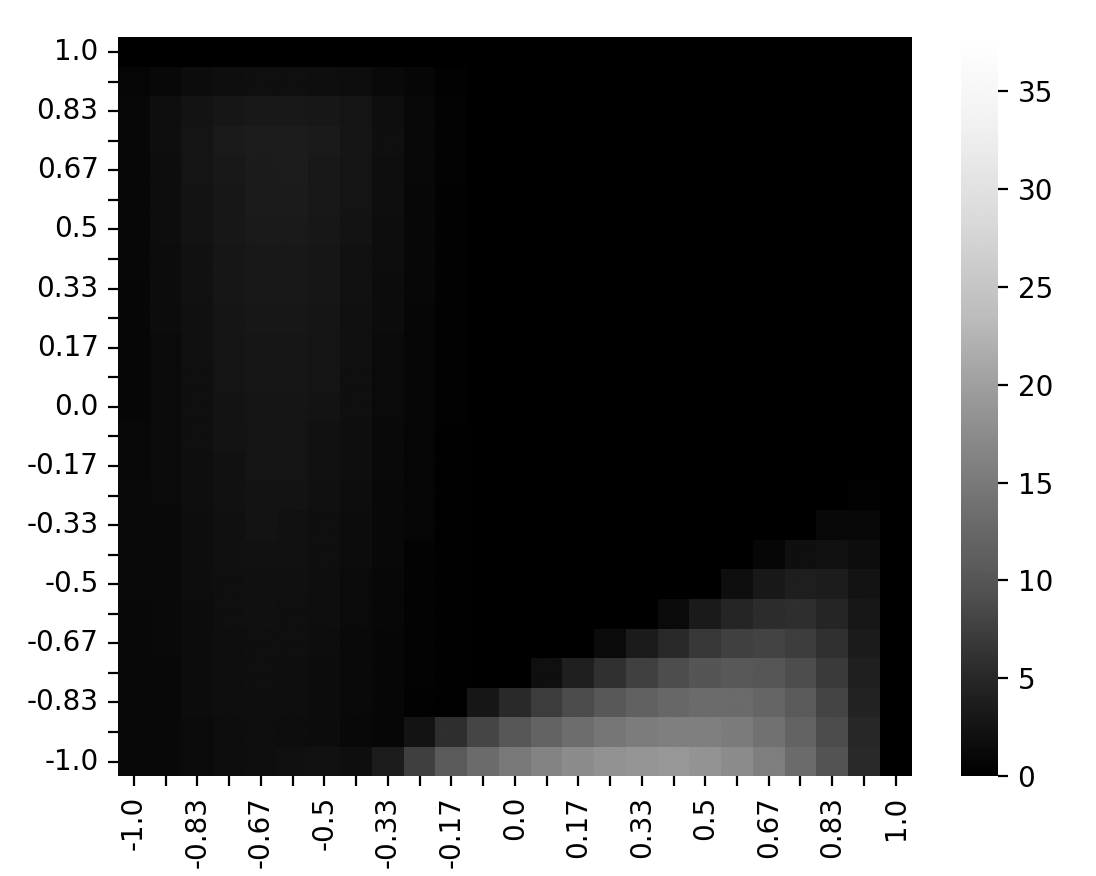}
	        \caption{Percent error between the learned and 
	        	optimal value on the training threat field.}
	        \label{fig:training_error}
	\end{subfigure}
	\begin{subfigure}{0.49\columnwidth}
		\centering
	        \includegraphics[width=\textwidth]{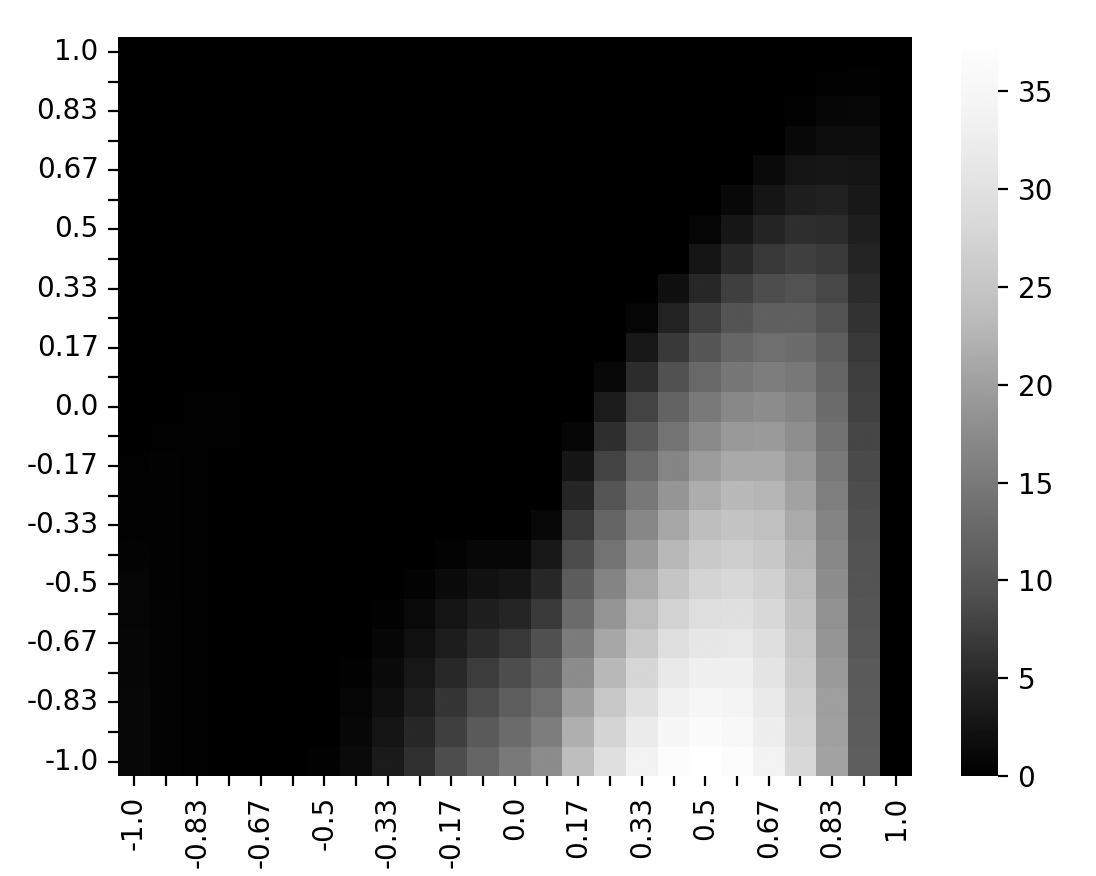}
	        \caption{Percent error between learned and optimal value
	        	for the test threat field.}
	        \label{fig:testing_error}
	\end{subfigure}
    \caption{Percent error for a threat field provided for training 
       	and for a different test threat field.}
    \label{fig:generalization-error}
\end{figure}

Figure~\ref{fig:training_error} shows the error between the learned and optimal
value functions for the training threat field. As expected, this error is small
for most initial states. Figure~\ref{fig:testing_error} shows the error between
the learned and optimal value functions for the test threat field. The error is
small in some regions, but there is a significant set of initial states where
the error is large ($\sim$30\%). The initial states where the learned policy
results in high error lie in the same regions of the environment for the
training and test fields. However, for the test threat field, the error has a
larger magnitude and extends farther from the edge of the threat field.

In summary, we note the potential for generalization of the proposed IRL model
to new threat fields unseen during training, especially if a diverse set of
threats is used during training.

\subsection{Policy Discrimination}

We return to the original motivation of performance analysis of an AV's
autonomous navigation from a limited amount of observed data. Here we focus on
the problem of the ability to the proposed IRL model to synthesize data that
match \emph{different} datasets with distinct characteristics. Informally,
suppose we have two different blackbox autonomy stacks that are characterized at
the high level by ``minimum threat exposure.'' The datasets of paths resulting
from each of these stacks under the same threat field are different, which
implies that the exact cost function used by each stack may be slightly
different. We would like to synthesize data that resembles each dataset.

\begin{figure}
	\centering
	\includegraphics[width=0.75\columnwidth]{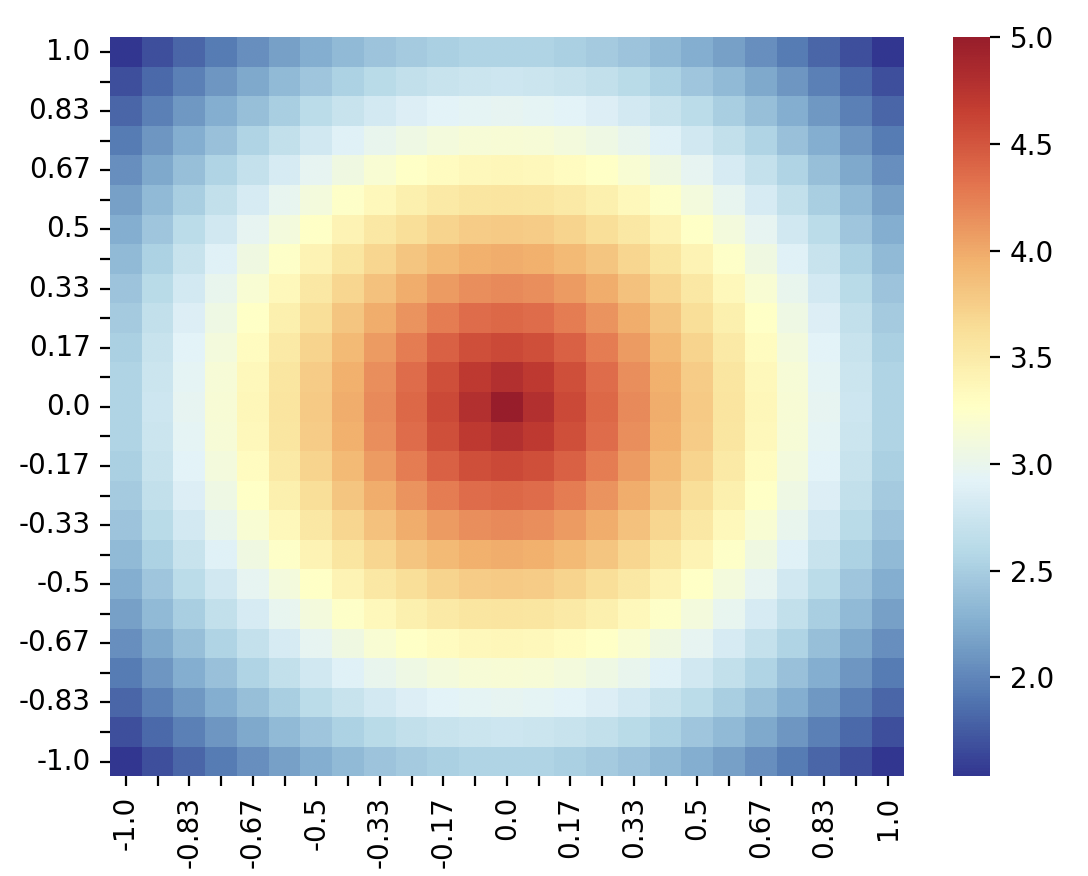}
	\caption{Simple threat field for the purposes of 
		training over two distinct datasets.}
	\label{fig:simple_threat}
\end{figure}

To this end, we consider two datasets $\trainingDataset\msub{A}$ and
$\trainingDataset\msub{B}$ for a static threat field shown in
Fig.~\ref{fig:simple_threat}. $\trainingDataset\msub{A}$ is similar to that
previously described, i.e., it consists of ``pure'' minimum threat exposure
paths.

For $\trainingDataset\msub{B},$ each path $\gridPath = (\mdpState_0,
\mdpState_1, \ldots, \mdpState_{\gridPathLength})$ minimizes $$\sum_{k =
	0}^{\gridPathLength} \threat(\xCoord[{\mdpState_k}]) + \left|
\vphantom{1^{2^3}}\xCoord[{\mdpState_k}][2] - \xCoordGoal[2]\right|$$ over all
paths between the same start and goal locations. Here
$\xCoord[{\mdpState_k}][2]$ indicates the second (vertical axis) coordinate,
which means that the cost function includes a $|\xCoord[{\mdpState_k}][2] -
\xCoordGoal[2]|$ is the vertical distance between $\xCoord[{\mdpState_k}]$ and
the goal location. Informally, due to this cost function, the paths in
$\trainingDataset\msub{B}$ go ``up and around'' the high threat area in the
center in Fig.~\ref{fig:simple_threat}, favoring early vertical movement towards
the goal. $\trainingDataset\msub{A}$ and $\trainingDataset\msub{B}$ consist of
$\nData = 500$ paths each.

\begin{figure}
	\centering
	\begin{subfigure}{0.9\columnwidth}
		\centering
		\includegraphics[width=\textwidth]{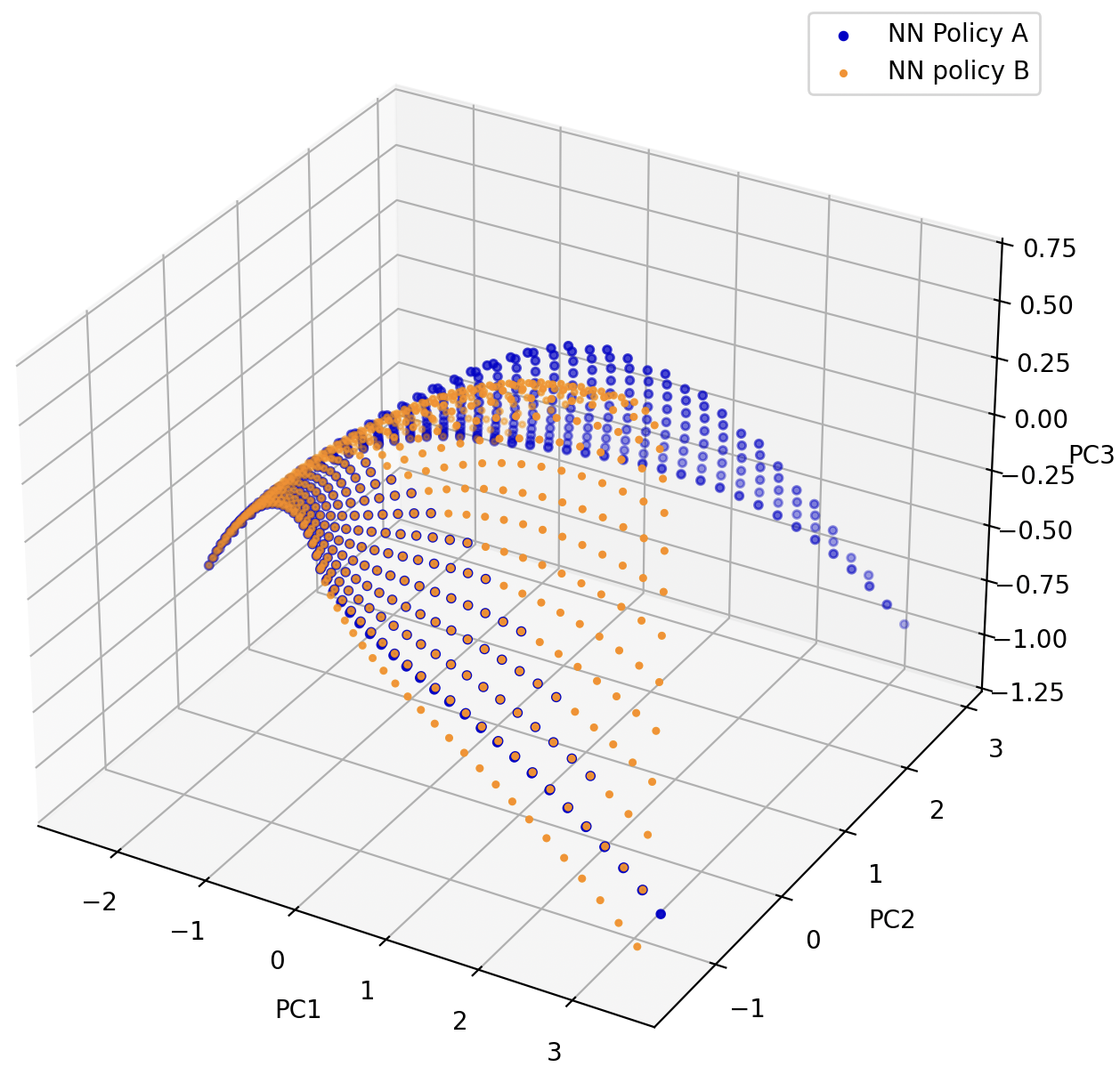}
		\caption{PCA view 1.}
		\label{fig:pca_view1}
	\end{subfigure}
	\begin{subfigure}{0.9\columnwidth}
		\centering
		\includegraphics[width=\textwidth]{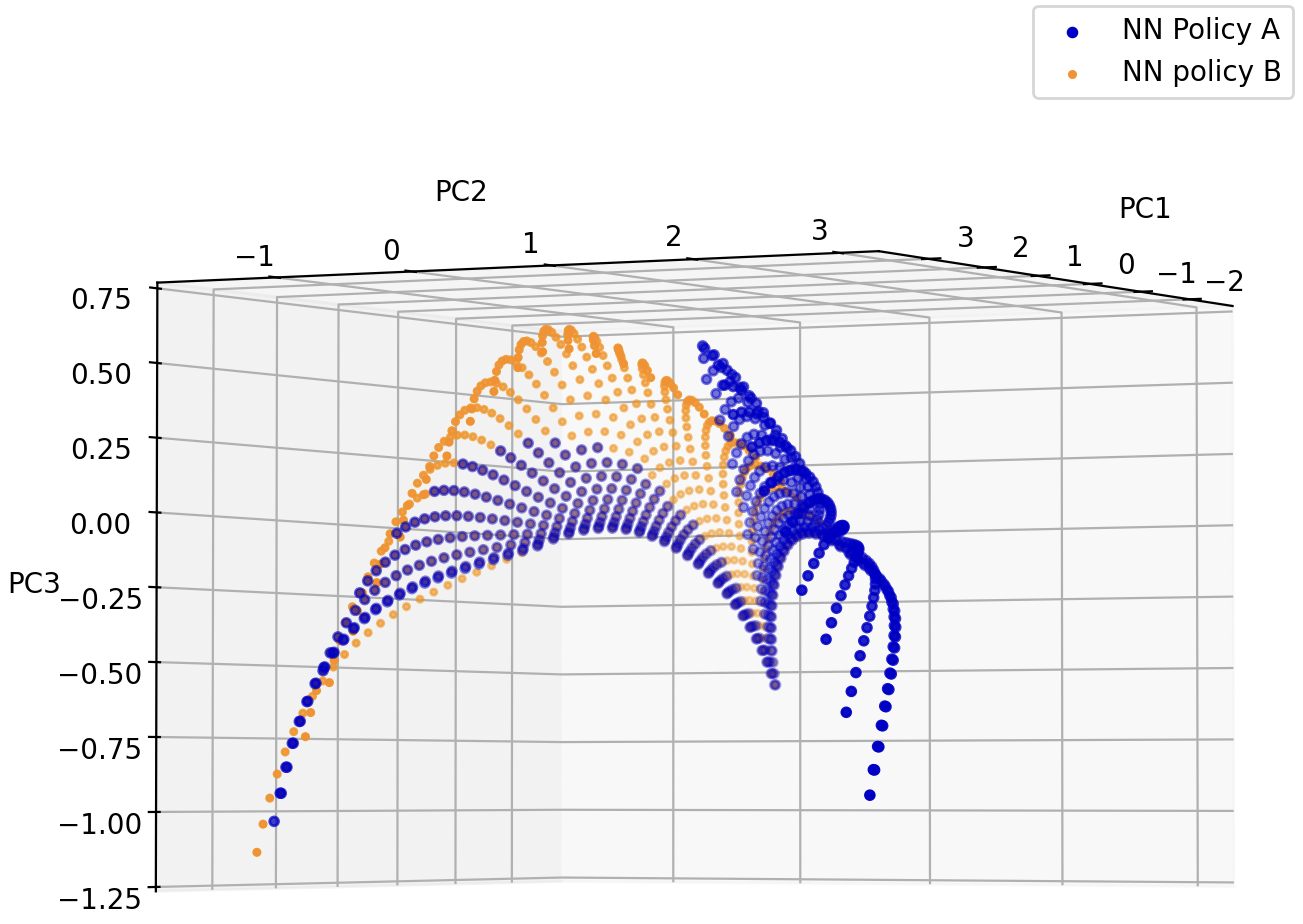}
		\caption{PCA view 2.}
		\label{fig:pca_view2}
	\end{subfigure}
	\caption{Scatter plot of first three principal components of output 
		datasets of the proposed IRL model trained under two distinct datasets.}
	\label{fig:pca_analysis}
\end{figure}

We train the IRL model separately over each of these two datasets, and then
synthesize $\nGen = 624$ paths (i.e., from every possible initial location
except the goal itself) from each of the learned policies.
{Note that we include additional entries in $r$ and $\phi$, as we split the distance
component in the original feature vector into $|\xCoord[{\mdpState_k}][1] -
\xCoordGoal[1]|$ and $|\xCoord[{\mdpState_k}][2] -
\xCoordGoal[2]|$, i.e. into a horizontal and vertical distance between $\xCoord[{\mdpState_k}]$ and the finish.}
Figure~\ref{fig:pca_analysis} shows a scatter plot of the first three principal
components of these synthesized outputs. The outputs of the IRL model trained on
$\trainingDataset\msub{A}$ are indicated in blue, whereas those trained on
$\trainingDataset\msub{B}$ are indicated in orange. Note that the two sets of
outputs have a subset of similar paths, but also a clear distinction in outputs,
as expected. These results signify that the output of the proposed IRL model is
distinct for policies generated under distinct datasets, allowing us to use the
output data to potentially develop more robust performance assessments or to
classify new paths according to relevant policies.

\section{Conclusions}
\label{sec-conclusions}

In this paper we proposed an inverse reinforcement learning model to synthesize
minimum threat exposure paths similar to a training dataset. 
We discussed details about the training process, including the IRL reward update
based on errors in a feature expectation vector. The Deep Q-learning model
used a probabilistic selection threshold for deciding on exploration or exploitation.
Via computational studies, we considered both
time-invariant (static) and time-varying (dynamic) threat fields. We found that
our proposed IRL model demonstrated excellent performance in synthesizing paths
from initial conditions not seen in the training dataset when the threat field
matched that used for training. Furthermore, we evaluated our model's
performance on unseen threat fields and found low error. Finally, we
demonstrated our model's ability to synthesize distinct datasets when we trained
it on different datasets with distinct characteristics.

\section*{Acknowledgements}
The authors are grateful to Prof. Randy Paffenroth, Mathematical Sciences Dept.
and Computer Science Dept. at WPI, for his guidance and insights during the
development of the proposed IRL model.

\addtolength{\textheight}{-13.5cm}   



%
%
%
%


\bibliography{references-noURL}

\end{document}